\documentclass[10pt,journal,compsoc]{IEEEtran}
\ifCLASSOPTIONcompsoc
\usepackage[nocompress]{cite}
\else
\usepackage{cite}
\fi

\ifCLASSINFOpdf
\else
\fi

\usepackage{amsmath}
\usepackage{algorithmic}
\usepackage{graphicx}
\usepackage[dvipsnames]{xcolor}
\usepackage{amssymb}
\usepackage{mathrsfs}
\usepackage{multirow}
\usepackage{array}
\usepackage{url}
\usepackage{wrapfig}
\usepackage{xspace}
\usepackage{booktabs}

\usepackage{bm}
\usepackage{siunitx}
\usepackage[super]{nth}

\usepackage[pagebackref=true,breaklinks=true,letterpaper=true,colorlinks,linkcolor=blue,citecolor=blue,bookmarks=false]{hyperref}

\usepackage{bbm}
\usepackage{colortbl}
\usepackage{lipsum}

\usepackage{colortbl} %
\definecolor{mygray-bg}{gray}{0.9}

\usepackage[short]{optidef}

\usepackage{svg}
\newcommand{\et}[2]{${#1}^{\pm{#2}}$}

\newcommand{\etr}[2]{$\boldsymbol{{#1}}^{\pm{#2}}$}
\newcommand{\etbb}[2]{$\underline{{#1}}^{\pm{#2}}$}

\definecolor{baselinecolor}{gray}{.9}

\usepackage[switch]{lineno}

\begin{document}

\title{EigenActor: Variant Body-Object Interaction Generation Evolved from \\Invariant Action Basis Reasoning}
	\author{Xuehao~Gao,~Yang~Yang,~Shaoyi~Du,~Yang~Wu, Yebin~Liu,~and~Guo-Jun~Qi%
}

\markboth{IEEE Transactions on Pattern Analysis and Machine Intelligence}{}


\IEEEtitleabstractindextext{
\begin{abstract}

This paper explores a cross-modality synthesis task that infers 3D human-object interactions (HOIs) from a given text-based instruction. Existing text-to-HOI synthesis methods mainly deploy a direct mapping from texts to object-specific 3D body motions, which may encounter a performance bottleneck since the huge cross-modality gap. In this paper, we observe that those HOI samples with the same interaction intention toward different targets, e.g., “lift a chair” and “lift a cup”, always encapsulate similar action-specific body motion patterns while characterizing different object-specific interaction styles. Thus, learning effective action-specific motion priors and object-specific interaction priors is crucial for a text-to-HOI model and dominates its performances on text-HOI semantic consistency and body-object interaction realism. In light of this, we propose a novel body pose generation strategy for the text-to-HOI task: infer object-agnostic canonical body action first and then enrich object-specific interaction styles. Specifically, the first canonical body action inference stage focuses on learning intra-class shareable body motion priors and mapping given text-based semantics to action-specific canonical 3D body motions. Then, in the object-specific interaction inference stage, we focus on object affordance learning and enrich object-specific interaction styles on an inferred action-specific body motion basis. Extensive experiments verify that our proposed text-to-HOI synthesis system significantly outperforms other SOTA methods on three large-scale datasets with better semantic consistency and interaction realism performances.


\end{abstract}

	\begin{IEEEkeywords}
	Text-driven Human-object Interaction Synthesis, Action-specific Motion Inference, Object-specific Interaction Reasoning.
\end{IEEEkeywords}}

\maketitle

\section{Introduction}

\label{sec:intro}
With the rapid gains in hardware and generative models, recent years have witnessed a significant breakthrough in AI-generated content (AIGC) \cite{cao2023comprehensive,zhang2023complete,zhang2023one,wu2023ai,li2023generative}. AIGC uses artificial intelligence technologies to assist or replace manual creation by generating content ($e.g.$, images \cite{reed2016generative,zhang2017stackgan}, videos \cite{blattmann2023align}, 3D meshes \cite{lin2023magic3d} and scenes \cite{yang2021text,gao2024multi}) based on user-inputted requirements. As a vital component of AIGC studies, data-driven human motion synthesis generates natural and realistic 3D body poses that can be used in wide-range applications, including virtual avatars, digital animation, human-robot interaction, game design, film script visualization, and AR/VR content creation \cite{zhu2023human,yang2022motion,guo2015adaptive,gao2021efficient}.

Recent human motion synthesis methods can be grouped into different specific generative sub-tasks based on their different condition input types, such as start-end positions \cite{zhao2023synthesizing}, movement trajectories \cite{karunratanakul2023guided,gao2023decompose}, scene contexts \cite{hassan2021stochastic,gao2024multi}, textual descriptions \cite{guo2022generating,petrovich2022temos,tevet2022human}, background music \cite{zhuang2022music2dance,li2021ai}, and speech audios \cite{qian2021speech,liu2022beat}. These diverse modalities of condition inputs reflect special requirements for human motion synthesis in different application contexts \cite{arikan2002interactive,guo2015adaptive,wang2022towards,gao2023learning}. However, most text-to-motion synthesis methods focus on planning body self-motions (e.g., \textit{walking} and \textit{running}) in isolation from object interaction contexts. However, many real-world human poses are manipulation-oriented and thus naturally harmonize with the 3D movements of the object they interact with. To fill this gap, manipulation-oriented human-object interaction (HOI) synthesis requires the generated virtual characters to operate a target 3D object as the user intended \cite{taheri2020grab,newbury2023deep,chao2021dexycb, li2023controllable}. For example, based on a given textual “\textit{take a photo with a camera}" command and a given 3D target \textit{camera} object, a text-to-HOI synthesis system would infer 3D full-body poses and 3D camera movements jointly and compose them into a realistic 3D \textit{photo-taking} animation. Therefore, without leaving human-object interaction behind, text-to-HOI generation focuses on object-specific manual manipulation and synthesizes desired body-object 3D co-movements based on a given text-based interaction intention. 

\begin{figure}[t]
	\centering
	\includegraphics[width=0.5\textwidth]{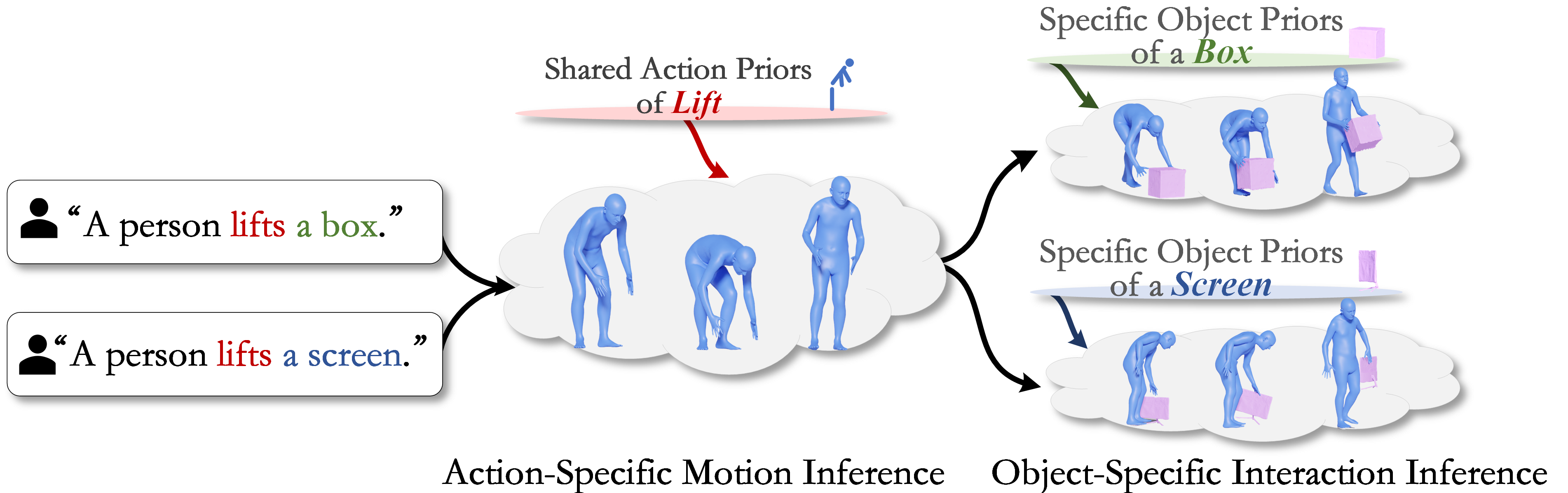}	
	\caption{In the text-to-HOI task, body motion reasoning can be factorized into two sequential stages: action-specific motion inference and object-specific interaction inference. The action-specific canonical body motion inferred from given textual instruction can serve as a primitive action basis for object-specific interaction reasoning.}
	\label{teaser}
\end{figure}

As a challenging cross-modality task, text-to-HOI synthesis requires a joint and holistic understanding of 3D body motions, 3D object motions, and natural languages \cite{bohg2013data,zhao2022learn}. Specifically, as shown in Fig. \ref{teaser}, human poses synthesized from a text-to-HOI generation system should be jointly harmonious with multiple conditional contexts: (1) \textit{Text-based Instruction}. A textual command defines a specific human-object interaction intention. Thus, the synthesized human-object interactions should reflect their intended manipulation semantics, thus exhibiting text-HOI semantic consistency; (2) \textit{Interaction Target}. Given a specific 3D object as an interaction target, a human pose should also harmonize with its shape, affordances, and functionality conditions, thus performing physically plausible manipulation. Therefore, synthesizing human motions in the text-to-HOI task is a challenging joint inference problem based on these text-object condition contexts. 

\begin{figure*}[t]
	\centering
	\includegraphics[width=1\textwidth]{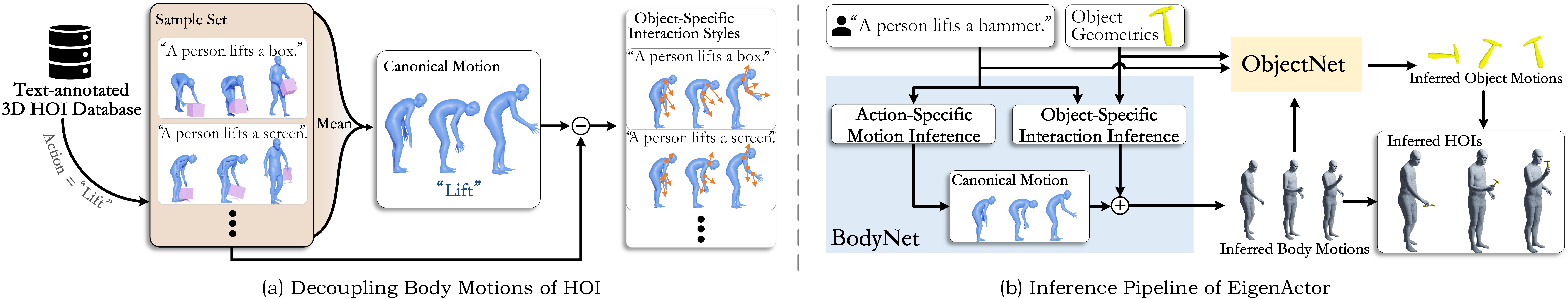}	
	\caption{Architecture Overview: (a) We first encapsulate an intra-class canonical pose sequence from category-specific diverse body motion samples. Then, we characterize object-specific interaction styles based on the evolution from actions-specific canonical poses to the body poses interacting with objects; (b) With factorized action-specific motion and object-specific interaction priors, 3D body poses inferred from EigenActor conform to the intended semantics and naturally interact with the object they manipulate.
		﻿
	}
	\label{pipeline}
\end{figure*}

Our core insight is that human poses in HOIs are composed of \textit{action-specific motion patterns} and \textit{object-specific interaction styles}, as shown in Fig. \ref{pipeline} (a). Those HOI samples with the same interaction intention semantics tend to reflect similar action-specific body motion patterns. For example, in \textit{drink with a bowl} and \textit{drink with a bottle} actions, their arms often encapsulate similar motion tendencies (raising up and bending). Thus, mapping a text-based action label to its intrinsic body motion patterns is a crucial step toward realistic text-to-HOI synthesis. Based on this shareable intra-class body motion knowledge, an object-agnostic canonical body action can be inferred from a given textual condition and serve as a primitive action basis for evolving object-specific interaction styles. 

In this paper, we propose a powerful text-to-HOI system named EigenActor to improve text-HOI semantic consistency and body-object interaction realism. Specifically, as shown in Fig. \ref{pipeline} (b), EigenActor has two core components: (1) BodyNet focuses on inferring a 3D full-body pose sequence from given text-object conditions and factorizes this task into action-specific canonical motion inference and object-specific interaction style inference stages. With the decoupled action-specific motion and object-specific interaction priors, human poses generated from BodyNet would not only conform to the intended semantics but also naturally interact with the objects they manipulate; (2) ObjectNet plans the 6-DoF postures of a target object on each frame based on multiple condition contexts, including inferred body poses, given text instruction and object shape. Benefiting from affordance-guided contact part inference and interaction optimization mechanisms, the inferred postures of a specific object target jointly harmonize with its affordances, functionality and physics.

We perform both qualitative and quantitative evaluations on popular text-to-HOI benchmarks, including HIMO \cite{lv2024himo}, FullBodyManipulation \cite{li2023object}, and GRAB \cite{taheri2020grab}. The comprehensive performance comparisons on these datasets validate that EigenActor significantly outperforms other text-to-HOI methods on three aspects: (1) \textit{Consistency between text-HOI semantics}. The HOI samples synthesized from EigenActor conform better to the intended interaction semantics; (2) \textit{Realism of body-object interactions}. EigenActor significantly improves realism and physical plausibility performances of inferred body-object interactions; (3) \textit{Robustness of Few-shot Learning}. Benefiting from shareable action-specific prior knowledge, EigenActor significantly improves the robustness and generalization performances of few-shot learning.

Overall, the main contributions in this paper are summarized as follows:
\begin{itemize}
	\item We propose a novel two-stage inference strategy for synthesizing realistic 3D body poses in the text-to-HOI task. With the decoupled action-specific motion and object-specific interaction priors, the body poses generated from given text-object conditions would conform to the intended semantics and naturally interact with the target object.  
	\item We develop a powerful object motion predictor that plans a plausible object posture sequence for the inferred body poses. Benefiting from  contact part inference and interaction optimization mechanisms, inferred object postures are harmonious with object affordances, functionality and physics. 
	\item Integrating effective body motion and object motion inference components, we develop a high-performance text-to-HOI synthesis system named EigenActor that outperforms state-of-the-art methods with better text-HOI semantic consistency and body-object interaction realism performances.
\end{itemize}

\section{Related Work}
\label{related_work}
\subsection{3D Human Motion Synthesis}
\label{t2m}
Human motion synthesis aims to generate natural human pose sequences as users intend, showing immense potential for real-world applications \cite{guo2015adaptive,zhu2023human}. Data-driven human motion synthesis has recently made fruitful attempts and attracted increasing attention. Human motion synthesis models can be grouped into different specific generative sub-tasks based on their different conditional input types, such as past motions \cite{li2018convolutional,mao2022weakly}, 3D scenes \cite{starke2019neural,hassan2021stochastic}, action labels \cite{guo2020action2motion,petrovich2021action}, textual descriptions \cite{ahuja2019language2pose,ghosh2021synthesis}, background music \cite{zhuang2022music2dance,li2021ai}, and speech audio \cite{qian2021speech,liu2022beat}. For example, SAMP \cite{hassan2021stochastic} focuses on scene-conditioned human motion synthesis and generates full-body motion following the inferred path and target. MotionGPT \cite{jiang2023motiongpt} proposes a text-to-motion generation system that synthesizes human motions from given textual descriptions. Zhang et al. \cite{zhang2024motiondiffuse} adopt a diffusion-based generation scheme and develop it into a text-driven human motion synthesis system named Motiondiffuse. Yiannakidis et al. \cite{aristidou2022rhythm} introduce a powerful music-driven human dance sequence system that generates realistic and diverse dance sequences conditioned on given music inputs. Wang et al. \cite{wang2019combining} learn generative networks from prerecorded human motion data and utilize them to generate natural-looking human motions consistent with various input constraints. Xu et al. \cite{xu2023smpler} propose a SMPL-based transformer framework named SMPLer to generate 3D human poses from a given monocular input image. These diverse conditional modalities reflect specific requirements for human motion synthesis in different application contexts, thus inspiring more human motion synthesis studies.


\subsection{Denoising Diffusion Probabilistic Model}
A series of generative models are introduced into the human motion synthesis task, including Variational Autoencoders (VAEs) \cite{hassan2021stochastic,wang2022humanise,guo2022generating,guo2022generating} and Generative Adversarial Networks (GANs) \cite{cui2021efficient,wang2019combining}. However, these attempts still suffer their potential limitations. Specifically, VAEs limit their learned distributions to normal latent distributions, thus enforcing strong prior assumptions on target human motion distributions. Without an effective training strategy, GANs tend to suffer from mode collapse and vanishing. Notably, diffusion models learn the target distribution by denoising its noised samples step by step \cite{NEURIPS2020_4c5bcfec}. Therefore, compared with these generative models, diffusion models are free from prior assumptions on the target human motion distribution, thus significantly facilitating the text-to-HOI task. For example, MDM \cite{tevet2022human} introduces a classifier-free diffusion-based generative model for the text-driven human motion task. MLD \cite{chen2023executing} develops motion diffusion into the latent space and learns the cross-modal mapping between text embedding and motion embedding. GUESS \cite{gao2024guess} develops a cascaded latent diffusion model that gradually enriches 3D human motions synthesized from given textual descriptions. CAMDM \cite{chen2024taming} proposes a transformer-based conditional autoregressive motion diffusion model, which takes the character’s historical motion as its inputs and can generate a range of diverse potential future motions conditioned on high-level, coarse user control. Encouraged by these fruitful attempts of diffusion models on various generation tasks, we adopt the diffusion model as our generative framework and thus develop a diffusion-based text-to-HOI synthesis system named EigenActor.  

\subsection{Text-driven Human-Object Interaction Generation}
Most 3D human motion synthesis methods focus on planning body self-motions in isolation from object interaction contexts. However, many real-world human poses are manipulation-oriented and naturally harmonize with the 3D movements of the object they interact with. To fill this gap, manipulation-oriented human-object interaction (HOI) synthesis requires the generated virtual characters to operate a target 3D object as the user intended \cite{bhatnagar2022behave,lv2024himo, yang2024f}. In this case, conditional HOI synthesis methods focus on inferring body-object 3D co-movements from given condition contexts, such as text-based instructions, object 3D movements, and HOI 3D motion histories. For example, OMOMO \cite{li2023object} generates a 3D human pose sequence conditioned on a given movement of the 3D object interacting with. GRAB \cite{taheri2020grab} proposes a powerful generative system that predicts 3D hand grasps for specified 3D object shapes, performing diverse in-hand manipulations. InterDiff \cite{xu2023interdiff} develops a diffusion-based generative model that predicts future human-object co-movements from their 3D interaction histories. Recently, the task of text-to-HOI synthesis is attracting more and more attention \cite{braun2024physically,ghosh2022imos,peng2023hoi}. Cha et al. \cite{cha2024text2hoi} develop a text-conditioned synthesis model that infers 3D hand-object interactions from textual instructions. CHOIS \cite{li2023controllable} plans 3D body-object co-movements from given textual controls. F-HOI \cite{yang2024f} plans 3D HOIs from given text-based fine-grained semantics. HIMO \cite{lv2024himo} proposes a large-scale MoCap dataset of full-body human interacting with multiple objects and develops a baseline model. CG-HOI \cite{diller2024cg} develops an approach to generate realistic 3D human-object interactions from a text description and given static object geometry to be interacted with. NIFTY \cite{kulkarni2024nifty} creates a neural interaction field attached to a specific object, which outputs the distance to the valid interaction manifold given a human pose as input.

Previous text-to-HOI synthesis works mainly adopt a one-stage body motion inference strategy that models a direct cross-modality mapping between text-based action descriptions and their 3D body poses. The key insight missing is that those HOI samples with the same interaction intention toward different targets always encapsulate similar action-specific body motion patterns while characterizing different object-specific interaction styles. Thus, with effective decoupled action-specific motion priors and object-specific interaction priors, the body poses inferred from given joint text-object conditions would conform to the intended semantics and naturally interact with the target object, thus significantly benefiting the text-to-HOI synthesis task.


\section{Problem Formulation}
In this paper, we introduce a powerful text-to-HOI synthesis system named EigenActor that infers a $N$-frame full-body pose sequence $\boldsymbol{b}_{1:N}$ and object movement sequence $\boldsymbol{o}_{1:N}$ from a specified 3D object geometry $\boldsymbol{g}$ and human-object interaction text $\boldsymbol{t}$ as: $\{\boldsymbol{b}_{1:N}, \boldsymbol{o}_{1:N}\}=\mathcal{F}(\boldsymbol{g},\boldsymbol{t})$. Specifically, at $n$-th frame, we represent its 3D body pose $\boldsymbol{b}_{n}$ with inferred SMPL parameters \cite{loper2015smpl,pavlakos2019expressive}, including root translation, axis-angle rotations for the body
joints, and face expression parameters. 6-DOF object posture $\boldsymbol{o}_{n}$ contains 3D rotation and 3D translation parameters of a object at $n$-th frame. Object geometry $\boldsymbol{g}$ represents the shape of a 3D object with $P$ down-sampled vertices. Textual instruction $\boldsymbol{t}$ describes an intended human-object interaction, such as “\textit{A person takes a picture with a camera}”. The performance of a text-HOI system $\mathcal{F}$ is reflected in the text-HOI semantic consistency and body-object interaction realism of its generated HOI samples.   
 
\begin{figure*}[t]
	\centering
	\includegraphics[width=0.85\textwidth]{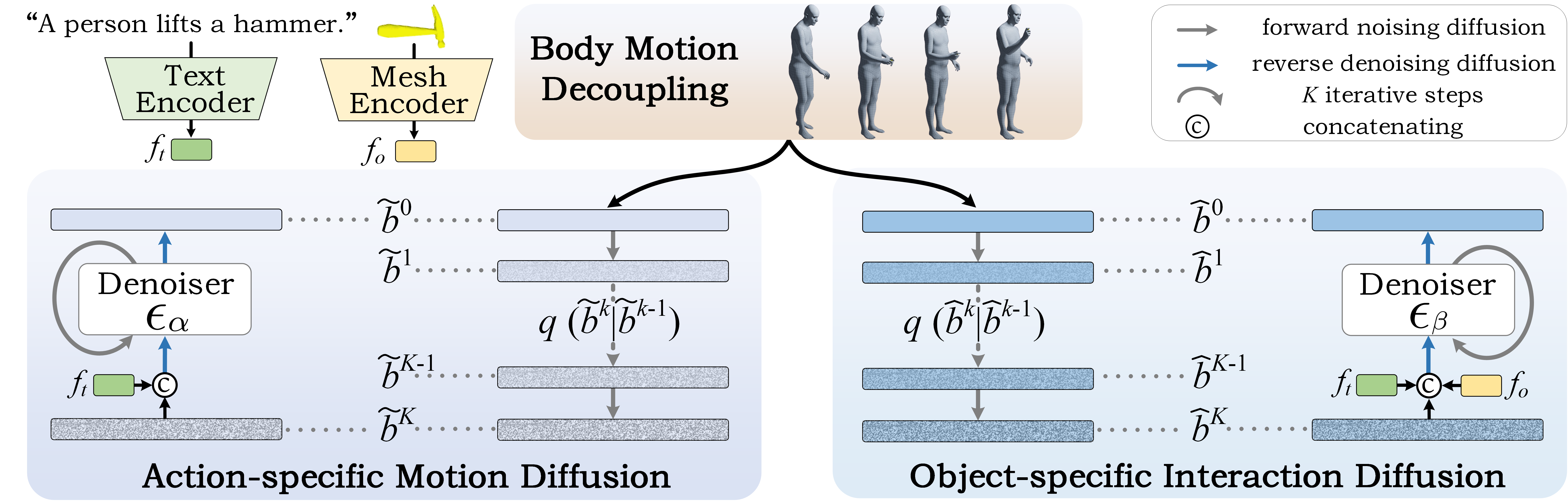}	
	\caption{BodyNet Module Overview. BodyNet factorizes the body pose reasoning task of text-to-HOI into two stages: synthesize action-specific canonical motion first and then enrich it with inferred object-specific interaction styles. With a denoising-based diffusion strategy, action-specific motion diffusion learns the conditional distribution from text-based intended semantics to its intra-class canonical 3D body motions. Object-specific interaction diffusion learns the conditional distribution from text-object joint conditions to body interaction styles.}
	\label{body_net}
\end{figure*} 

\section{Methodology}
To construct our text-to-HOI generation system EigenActor, as shown in Fig. \ref{pipeline} (b), we consider two core components: \textit{BodyNet} that synthesizes 3D full-body poses from given text-based instruction and specific object shape condition contexts; \textit{ObjectNet} that plans 3D object motions conformed with inferred body poses and given text-object conditions. In the following, we elaborate on their technical details.

\subsection{\textit{BodyNet}: \normalsize{Synthesize Full-Body 3D Poses}}
Instead of modeling body motion directly, the key insight missing is that human poses in HOIs intertwine action-specific motion patterns and object-specific interaction styles. Thus, as shown in Fig. \ref{body_net}, to improve text-body semantic consistency and object-body interaction realism, BodyNet proposes a novel two-stage body pose inference strategy: synthesize action-specific canonical motion first and then enrich it with inferred object-specific interaction styles. 
\subsubsection{Body Motion Decoupling}
In the pose motion decoupling stage, we factorize action-specific motion priors and object-specific interaction priors from text-annotated HOI samples. Specifically, based on annotated textual descriptions, we group body pose samples into different action sets. The 3D body poses in the same action set have the same intention semantics, such as “\textit{kick}”, “\textit{pick up}”, “\textit{eat}”. Intuitively, diverse samples within the same action set tend to encapsulate similar intra-class motion patterns. Then, as for each action set, we average its all body pose samples frame by frame and thus obtain an action-specific 3D mean-pose sequence. Finally, we abstract a canonical action set $\mathcal{X}$ from a text-annotated HOI dataset. Each element of $\mathcal{X}$ is a $N$-frame body pose sequence $\boldsymbol{\widetilde{b}}_{1:N}$ that reflects generic motion patterns of a specific action semantic. 

Given an arbitrary body pose sequence $\boldsymbol{b}_{1:N}$, its object-specific interaction styles is characterized by the evolution from its action-specific canonical poses $\boldsymbol{\widetilde{b}}_{1:N}^{i}$ to $\boldsymbol{b}_{1:N}$. $\boldsymbol{\widetilde{b}}_{1:N}^{i}$ is a canonical action sample retrieved from $\mathcal{X}$ based on the action category of $\boldsymbol{b}_{1:N}$. For simplicity, we omit superscript $i$ when $i$ represents an arbitrary action category index, \textit{e.g.}, taking $\boldsymbol{\widetilde{b}}_{1:N}^{i}$ as $\boldsymbol{\widetilde{b}}_{1:N}$. Therefore, we formulate the object-specific interaction style of a body pose sequence $\boldsymbol{b}_{1:N}$ as:
\begin{equation}
	\boldsymbol{\widehat{b}}_{n} =  \boldsymbol{b}_{n} - \boldsymbol{\widetilde{b}}_{n}, \; \text{where} \; n \in \{1, \dots, N\}.
\end{equation}
   
By repeating these operations on each 3D body motion sample $\boldsymbol{b}_{1:N}$, we finally factorize it into action-specific motion patterns $\boldsymbol{\widetilde{b}}_{1:N}$ and object-specific interaction styles $\boldsymbol{\widehat{b}}_{1:N}$. For simplicity, we omit their subscripts $1:N$ when they represents an arbitrary $N$-frame sequence, e.g., taking $\boldsymbol{\widetilde{b}}_{1:N}$ as $\boldsymbol{\widetilde{b}}$ and $\boldsymbol{\widehat{b}}_{1:N}$ as $\boldsymbol{\widehat{b}}$.

\subsubsection{Action-specific Motion Diffusion}
Given a textual instruction condition $\boldsymbol{t}$, action-specific motion inference module adopts a diffusion-based generative model to learn the cross-modality mapping from $\boldsymbol{t}$ to its corresponding intra-class canonical motions $\boldsymbol{\widetilde{b}}$. Specifically, the denoising diffusion model \cite{sohl2015deep} learns the posterior distribution $p(\boldsymbol{\widetilde{b}}|\boldsymbol{t})$ with the noising-denoising strategy. In the forward noising diffusion stage, we inject a random noise signal into the ground-truth $\boldsymbol{\widetilde{b}}$. Then, in the reverse iterative denoising stage, we develop a Transformer-based denoiser to iteratively denoise noised $\boldsymbol{\widetilde{b}}$ based on $\boldsymbol{t}$.

Inspired by the stochastic diffusion process in Thermodynamics, the forward noising diffusion of is $\boldsymbol{\widetilde{b}}$ modeled as a $K$-step Markov process. Specifically, at the $k$-th noising step, a sampled Gaussian noise signal is injected into $\boldsymbol{\widetilde{b}}^{k-1}$ as:
\begin{equation}
	q\left(\boldsymbol{\widetilde{b}}^{k} \mid \boldsymbol{\widetilde{b}}^{k-1}\right)=\mathcal{N}\left(\sqrt{\alpha_k}\boldsymbol{\widetilde{b}}^{k-1},\left(1-\alpha_k\right) I\right),
\end{equation}
where the constant $\alpha_k \in (0,1)$ is a hyper-parameters for sampling. After a $K$-step forward noising stage, $\boldsymbol{\widetilde{b}}$ is noised into a sample sequence $\{\boldsymbol{\widetilde{b}}^{k}\}^{K}_{k=0}$. If $K$ is sufficiently large, $\boldsymbol{\widetilde{b}}^{K}$ will approximate a normal Gaussian noise signal.

In the following reverse denoising stage, we first deploy a pre-trained CLIP \cite{radford2021learning} as a text feature extractor to embed the given text condition $\boldsymbol{t}$ into its latent representation $\boldsymbol{f}_{t}$. Then, we develop a Transformer-based denoiser $\epsilon_\alpha$ that iteratively anneals the noise of $\{\boldsymbol{\widetilde{b}}^{k}\}^{K}_{k=0}$ and reconstruct $\boldsymbol{\widetilde{b}}$ conditioned on $\boldsymbol{f}_{t}$. Specifically, $\epsilon_\alpha(\boldsymbol{\widetilde{b}}^{k}, k, \boldsymbol{f}_{t})$ learns the conditional distribution $p(\boldsymbol{\widetilde{b}}|\boldsymbol{t})$ at the $k$-th denoising step as:
\begin{equation}
	\boldsymbol{\widetilde{b}}^{k-1} = \frac{1}{\sqrt{\alpha_k}}\boldsymbol{\widetilde{b}}^{k} -\sqrt{\frac{1}{\alpha_k}-1}\epsilon_\alpha(\boldsymbol{\widetilde{b}}^{k}, k, \boldsymbol{f}_{t}).
\end{equation}

In the training stage of the action-specific motion inference module, we optimize the denoiser $\epsilon_\alpha$ by minimizing its denoising error in the noising-denoising process as: 
\begin{equation}
	\begin{aligned}
		\mathcal{L}_{\widetilde{b}} = \mathbb{E}_{\epsilon, k}\left[\left\|\epsilon-\epsilon_\alpha\left(\boldsymbol{\widetilde{b}}^{k}, k, \boldsymbol{f}_{t}\right)\right\|_2^2\right],
	\end{aligned}
\end{equation}  
where $\epsilon \sim \mathcal{N}(0,1)$. In the inference stage, based on the given text condition embedding $\boldsymbol{f}_t$, denoiser $\epsilon_\alpha$ recursively denoises a sampled Gaussian noise signal and outputs the inferred action-specific motions $\boldsymbol{\widetilde{b}}^{\prime}$.

\subsubsection{Object-specific Interaction Diffusion}
Human poses in HOIs should not only be consistent with the intended semantics but also harmonize with the specific shape of the object they interact with. Thus, different shapes of interaction targets significantly enrich the intra-class diversity of body poses. Given a textual instruction $\boldsymbol{t}$ and specific object geometry $\boldsymbol{g}$, object-specific interaction style inference adopts a diffusion model to learn the posterior distribution $p(\boldsymbol{\widehat{b}}|\boldsymbol{t},\boldsymbol{g})$.

In the forward noising diffusion stage, ground-truth $\boldsymbol{\widehat{b}}$ is noised in a $K$-step Markov process. At the $k$-th noising step, a sampled Gaussian noise signal is injected into $\boldsymbol{\widehat{b}}^{k-1}$ as:
\begin{equation}
	q\left(\boldsymbol{\widehat{b}}^{k} \mid \boldsymbol{\widehat{b}}^{k-1}\right)=\mathcal{N}\left(\sqrt{\beta_k}\boldsymbol{\widehat{b}}^{k-1},\left(1-\beta_k\right) I\right),
\end{equation}
where $\beta_k \in (0,1)$ is a hyper-parameters for sampling. After a $K$-step forward noising stage, we obtain a noised sequence $\{\boldsymbol{\widehat{b}}^{k}\}^{K}_{k=0}$. In the following reverse denoising stage, we first deploy a pre-trained PointTransformer \cite{zhao2021point} as a object feature extractor to embed the given object geometry condition $\boldsymbol{g}$ into its latent representation $\boldsymbol{f}_{o}$. Then, we develop a Transformer-based denoiser $\epsilon_\beta$ to iteratively anneal the noise of $\{\boldsymbol{\widehat{b}}^{k}\}^{K}_{k=0}$ and reconstruct $\boldsymbol{\widehat{b}}$ conditioned on $\boldsymbol{f}_{t}$ and $\boldsymbol{f}_o$. Specifically, $\epsilon_\beta(\boldsymbol{\widehat{b}}^{k}, k, \boldsymbol{f}_{t}, \boldsymbol{f}_o)$ learns the conditional distribution $p(\boldsymbol{\widehat{b}}|\boldsymbol{t},\boldsymbol{g})$ at the $k$-th denoising step as:
\begin{equation}
	\boldsymbol{\widehat{b}}^{k-1} = \frac{1}{\sqrt{\beta_k}}\boldsymbol{\widehat{b}}^{k} -\sqrt{\frac{1}{\beta_k}-1}\epsilon_\beta(\boldsymbol{\widehat{b}}^{k}, k, \boldsymbol{f}_{t}, \boldsymbol{f}_o).
\end{equation}

In the training stage of the object-specific interaction inference module, we optimize the denoiser $\epsilon_\beta$ by minimizing its denoising error in the noising-denoising process as: 
\begin{equation}
	\begin{aligned}
		\mathcal{L}_{\widehat{b}} = \mathbb{E}_{\epsilon, k}\left[\left\|\epsilon-\epsilon_\beta\left(\boldsymbol{\widehat{b}}^{k}, k, \boldsymbol{f}_{t}, \boldsymbol{f}_o\right)\right\|_2^2\right],
	\end{aligned}
\end{equation}  
where $\epsilon \sim \mathcal{N}(0,1)$. In the inference stage, based on the given text and object condition embeddings (\textit{i.e.}, $\boldsymbol{f}_t$ and $\boldsymbol{f}_o$), denoiser $\epsilon_\beta$ denoises a sampled Gaussian noise signal to infer the object-specific interaction style $\boldsymbol{\widehat{b}}^{\prime}$. Then, based on the inferred action-specific motion basis $\boldsymbol{\widetilde{b}}^{\prime}$ and object-specific interaction styles $\boldsymbol{\widehat{b}}^{\prime}$, we can integrate them into a synthetic 3D body pose sequence $\boldsymbol{b}^{\prime}$ that both conforms to the intended semantics and harmonizes with specified object shapes as:
\begin{equation}
	\boldsymbol{{b}}_{n}^{\prime} =  \boldsymbol{\widetilde{b}}_{n}^{\prime} + \boldsymbol{\widehat{b}}_{n}^{\prime}, \; \text{where} \; n \in \{1, \dots, N\}.
\end{equation}

\begin{figure*}[t]
	\centering
	\includegraphics[width=0.9\textwidth]{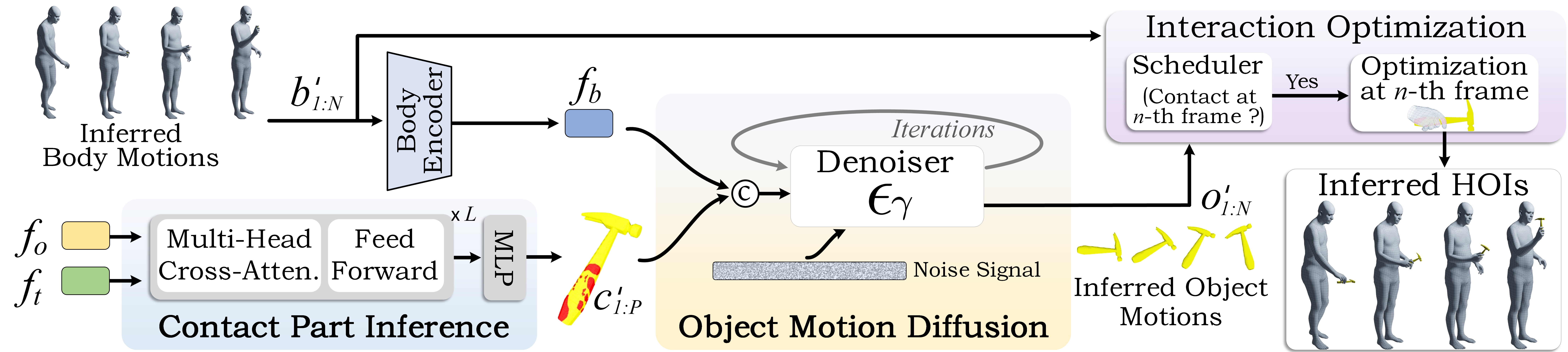}	
	\caption{ObjectNet Module Overview. ObjectNet contains three components: contact part inference, object motion diffusion, and hand-object interaction optimization. Contact part inference analyzes object-specific hand-contactable parts for the following object-hand interaction planning. Object motion diffusion infers 3D object movements from inferred body poses and contact parts. Interaction optimization integrates inferred 3D body-object co-movements and improves the realism of the manipulation between them.}
	\label{object_net}
\end{figure*}

\subsection{\textit{ObjectNet}: \normalsize{Plan Object 3D Movements}}
ObjectNet focuses on planning a $N$-frame object movement sequence $\boldsymbol{o}_{1:N}^{\prime}$ based on multiple condition contexts, including inferred body poses $\boldsymbol{b}^{\prime}_{1:N}$, text feature $\boldsymbol{f}_t$, and object feature $\boldsymbol{f}_o$. To construct the ObjectNet, as shown in Fig. \ref{object_net}, we consider three basic components: a contact part inference module, an object motion diffusion module, and an interaction optimization module.   

\subsubsection{Contact Part Inference}
Given a specified target object, analyzing its correct hand-contactable parts from its functional affordance is a foundational step toward reasoning realistic hand-object interactions. We represent hand-object contact using the distance between hand joints and object surface points. Specifically, based on ground-truth $\boldsymbol{b}_{1:N}$ and $\boldsymbol{o}_{1:N}$ parameters, we first retrieve a $N$-frame 3D body hand vertex sequence $\boldsymbol{v}_h\in \mathbb{R}^{N\times J\times3}$ and object vertex sequence $\boldsymbol{v}_o\in \mathbb{R}^{N\times P\times3}$, where $J$ and $P$ denote the number of hand joints and object points, respectively. Then, we compute the per-vertex distance map $\boldsymbol{d} \in \mathbb{R}^{N\times J\times P}$ at all $N$ frames, where each entry $\boldsymbol{d}_{n,j,p}$ encodes the $\ell_2$ distance between the $j$-th hand joint and $p$-th object point at the $n$-th frame.

In the distance map $\boldsymbol{d}$, the object points with less hand-contactable attributes have a higher value. Thus, we further normalize the original distance map $\boldsymbol{d}$ as:
\begin{equation}
	\boldsymbol{\overline{d}}_{n,j,p}=e^{-\frac{1}{2} \frac{\boldsymbol{d}_{n,j,p}^2}{\sigma^2}},
\end{equation}
where the constant $\sigma$ is the normalizing factor. In $\boldsymbol{\overline{d}}$, those object points that are closer to hand joints have a value closer to 1. Then, for an arbitrary $p$-th object point, we define its hand-contactable confidence based on the maximum value of the matrix $\boldsymbol{\overline{d}}_{1:N,1:J,p} \in \mathbb{R}^{N\times J}$. Thus, the ground-truth of object contact part inference is defined as:
\begin{equation}
	\boldsymbol{c}_{p}= \begin{cases}{1} & \text{ if } \operatorname{max}(\boldsymbol{\overline{d}}_{1:N,1:J,p})>\lambda \\ 0 & \text { otherwise },\end{cases}, \text{where} \; p \in \{1,\dots,P\}. 
\end{equation}
 
In the contact part inference module, we focus on developing a Transformer-based predictor $\mathcal{F}_{c}(\cdot)$ to infer object-specific hand-contactable parts $\boldsymbol{c}^{\prime}=\mathcal{F}_{c}(\boldsymbol{f}_o,\boldsymbol{f}_t)$ to approximate the ground-truth $\boldsymbol{c}$. Specifically, given $\boldsymbol{f}_o$ and $\boldsymbol{f}_t$, we feed them to a $L$-layer cross-attention Transformer for feature fusion. For simplicity, we introduce the technique we use in each layer. 

Firstly, a Multi-Head Attention is used for fusion feature embeddings as:
\begin{equation}
	\begin{aligned}
		& \operatorname{MultiHead}(\boldsymbol{Q}, \boldsymbol{K}, \boldsymbol{V})=\left[\operatorname{head}_1 ; \ldots ; \operatorname{head}_h\right] W^O \\
		& \text {where} \operatorname{head}_i=\operatorname{softmax}\left(\frac{\boldsymbol{Q}^i\left(\boldsymbol{K}^i\right)^T}{\sqrt{d_K}}\right) \boldsymbol{V}^i, 
	\end{aligned}
\end{equation}
where $W^{O}$ is a projection parameter matrix, $d_K$ is the dimension of the key $\boldsymbol{K}$ and $h$ is the number of the heads we use. We adopt a cross-attention strategy and get the query $\boldsymbol{Q}^{i}$, key $\boldsymbol{K}^i$, and value $\boldsymbol{V}^i$ from $\boldsymbol{f}_t$ and $\boldsymbol{f}_o$ for $i$-th head as:
\begin{equation}
	\boldsymbol{Q}^{i}= \boldsymbol{f}_t W^i_Q, \; \boldsymbol{K}^{i}= \boldsymbol{f}_o W^i_K, \; \boldsymbol{V}^{i}= \boldsymbol{f}_o W^i_V,
\end{equation}
where $W_Q^i$, $W_K^i$, and $W_V^i$ are projection parameter matrices. We then employ a residual connection and the layer normalization techniques to our architecture. We further apply a feed forward layer, again followed by a residual connection and a layer normalization following \cite{vaswani2017attention}. With $L$ such layers, we obtain a updated feature embedding and then feed it to a linear layer to predict a $P$-dimensional contact map $\boldsymbol{c}^{\prime}$, where $\boldsymbol{c}^{\prime}_{p}$ denotes the inferred hand-contactable probability of the $p$-th object point. In the training stage of the contact part inference module, all learnable parameters are optimized by minimizing the prediction error $\mathcal{L}_c$ as:
\begin{equation}
	\mathcal{L}_{c}= -\sum_{p=1}^P\left[\boldsymbol{c}_p \ln \boldsymbol{c}_p^{\prime}+\left(1-\boldsymbol{c}_p\right) \ln \left(1-\boldsymbol{c}_p^{\prime}\right)\right]. 
\end{equation}

\subsubsection{Object Motion Diffusion}
Given inferred $N$-frame body poses $\boldsymbol{b}^{\prime}_{1:N}$ and inferred $P$-point object contact map $\boldsymbol{c}_{1:P}^{\prime}$, object motion diffusion learns the posterior distribution $p(\boldsymbol{o}|\boldsymbol{b},\boldsymbol{c})$ with a noising-denoising strategy. In the forward noising stage, ground-truth $\boldsymbol{o}_{1:N}$ is noised in a $K$-step Markov process. At the $k$-th noising step, a sampled Gaussian noise signal is injected into $\boldsymbol{o}^{k-1}$ as:
\begin{equation}
	q\left(\boldsymbol{o}^{k} \mid \boldsymbol{o}^{k-1}\right)=\mathcal{N}\left(\sqrt{\alpha_k}\boldsymbol{o}^{k-1},\left(1-\alpha_k\right) I\right).
\end{equation}
After a $K$-step forward noising stage, we obtain a noised sequence $\{\boldsymbol{o}^{k}\}^{K}_{k=0}$. In the following reverse denoising stage, we first deploy a Transformer-based body motion feature extractor to embed $\boldsymbol{b}^{\prime}_{1:N}$ into its latent representation $\boldsymbol{f}_b$. A series of linear layers also embed $\boldsymbol{c}_{1:P}^{\prime}$ into $\boldsymbol{f}_c$. Then, we develop a Transformer-based denoiser $\epsilon_\gamma$ that iteratively anneals the noise of $\{\boldsymbol{o}^{k}\}^{K}_{k=0}$ and reconstruct $\boldsymbol{o}$ conditioned on $\boldsymbol{f}_{b}$ and $\boldsymbol{f}_c$. Specifically, $\epsilon_\gamma(\boldsymbol{o}^{k}, k, \boldsymbol{f}_{b}, \boldsymbol{f}_c)$ learns the conditional distribution $p(\boldsymbol{o}|\boldsymbol{b},\boldsymbol{c})$ at the $k$-th denoising step as:
\begin{equation}
	\boldsymbol{o}^{k-1} = \frac{1}{\sqrt{\gamma_k}}\boldsymbol{o}^{k} -\sqrt{\frac{1}{\gamma_k}-1}\epsilon_\gamma(\boldsymbol{o}^{k}, k, \boldsymbol{f}_{b}, \boldsymbol{f}_c).
\end{equation}

In the training stage of object motion diffusion, we optimize the denoiser $\epsilon_\gamma$ by minimizing its denoising error in the noising-denoising process as: 
\begin{equation}
	\begin{aligned}
		\mathcal{L}_{o} = \mathbb{E}_{\epsilon, k}\left[\left\|\epsilon-\epsilon_\gamma\left(\boldsymbol{o}^{k}, k, \boldsymbol{f}_{b}, \boldsymbol{f}_c\right)\right\|_2^2\right],
	\end{aligned}
\end{equation}  
where $\epsilon \sim \mathcal{N}(0,1)$. In the inference stage, based on the inferred body pose and object contact part condition embeddings (\textit{i.e.}, $\boldsymbol{f}_b$ and $\boldsymbol{f}_c$), denoiser $\epsilon_\gamma$ recursively denoises a sampled Gaussian noise signal to infer a $N$-frame 3D object motion sequence $\boldsymbol{o}^{\prime}_{1:N}$.

\subsubsection{Hand-Object Interaction Optimization} 
Based on the inferred $N$-frame object posture and body pose sequences ($i.e.$, $\boldsymbol{o}^{\prime}_{1:N}, \boldsymbol{b}^{\prime}_{1:N}$), we develop an interaction optimizer that enforces interaction constraints to improve the in-hand manipulation realism between them. Considering that hands and objects may not be in contact consistently across all $N$-frames, the interaction optimizer thus first dynamically detects the per-frame hand-object contact state. We then apply interaction constraints at those hand-object in-contact moments.

Specifically, we first retrieve $N$-frame 3D body and object vertex sequences using $\boldsymbol{o}^{\prime}_{1:N}$ and $\boldsymbol{b}^{\prime}_{1:N}$ parameters. Then, we calculate the inter-vertex distance matrix $\boldsymbol{d}^{\prime} \in \mathbb{R}^{N\times J\times P}$ between $J$ hand joints and $P$ object points. At an arbitrary $n$-th frame, if the minimum of its hand-object distance matrix $\boldsymbol{d}_{n, 1:J, 1:P}^{\prime}$ is smaller that a defined threshold $\delta$, we consider that hands and object are in contact at this moment and then calculate the in-hand manipulation error at the $n$-th frame as:
\begin{equation}
	\begin{aligned}
		I_{n} = ||\boldsymbol{d}_{n}^{\prime}-\boldsymbol{d}_{m}^{\prime}||_2 + ||\boldsymbol{d}_{n, j, p}^{\prime}||_2, \; \textit{where} \; \boldsymbol{d}_{n, j, p}^{\prime} \leq \delta.
	\end{aligned}
\end{equation}
Intuitively, the first item of ${I}_n$ defines a temporal consistency constraint that ensures the hand-object inter-vertex distances at an arbitrary in-contact frame ($i.e.$, $\boldsymbol{d}_{n}^{\prime}$) to keep the same as they are at the first in-contact moment ($i.e.$, $\boldsymbol{d}_{m}^{\prime}$). The second item of ${I}_n$ forces the distance between the in-contact vertex pairs to be zero. We calculate these in-hand manipulation errors at each hand-object in-contact frame and sum them into $\mathcal{L}_{I}$. We only apply the interaction error $\mathcal{L}_{I}$ in the last 10 denoising steps of object motion diffusion since the prediction in the early steps is extremely noisy.  

\section{Experiments}
In this section, we elaborate on the implementation details, the evaluation datasets, and the evaluation metrics. Then, we report extensive experimental results with ablation studies to show the effectiveness of EigenActor in these benchmarks.
\subsection{Implementation Details}
In the BodyNet, denoiser $\epsilon_\alpha$ and $\epsilon_\beta$ are both $8$-layer Transformers with $4$ heads. The number of action-specific and object-specific diffusion steps are both $1$k during training while $50$ during interfering. In the ObjectNet, contact part inference module deploys a $6$-layer cross-attention Transformer. Body feature extractor is a $4$-layer Transformers with $4$ heads. Denoiser $\epsilon_\gamma$ is also a $8$-layer Transformers with $4$ heads. The number of object motion diffusion steps is $1$k during training while $50$ during interfering. The dimensions of latent representations are set to $\boldsymbol{f}_b\in\mathbb{R}^{512}$, $\boldsymbol{f}_t\in\mathbb{R}^{1256}$, $\boldsymbol{f}_o\in\mathbb{R}^{512}$, and $\boldsymbol{f}_c\in\mathbb{R}^{512}$, respectively. The training of EigenActor consists of four stages: action-specific motion diffusion training (Stage-I), object-specific interaction diffusion training (Stage-II), contact part inference training (Stage-III), and object motion diffusion training (Stage-IV). The batch sizes is set to $128$ during Stage-I, II, and IV and set to 64 during Stage-III. The training epoch is set to $6$K during Stage-I, II, and IV and set to $3$K during Stage-III. In all four training stages, their optimizers are both Adam \cite{kingma2014adam} with a fixed learning rate of $10^{-4}$. The training/inference of EigenActor is implemented with PyTorch 1.11 on four RTX-3090 GPUs. 


\subsection{Dataset}
We adopt three common-used and popular text-to-HOI synthesis benchmarks to verify the effectiveness of EigenActor, including HIMO \cite{lv2024himo}, FullBodyManipulation \cite{li2023object}, and GRAB \cite{taheri2020grab}. In the following, we briefly introduce these large-scale datasets.

\textbf{HIMO} \cite{lv2024himo} is a text-annotated 3D HOI dataset that comprises diverse body-object interaction samples and their textual descriptions. 34 subjects participated in the collection of HIMO, resulting in 3,376 HOI sequences, 9.44 hours and 4.08M frames. HIMO chooses 53 common household objects in various scenes including the dining rooms, kitchens, living rooms and studies from ContactDB \cite{brahmbhatt2019contactdb} and Sketchfab \cite{jacobs2022sketchup}. HIMO elaborately annotates a fine-grained textual description of each 3D HOI sequence. We follow the official protocol of HIMO and split train, test, and validation sets with the ratio of 0.8, 0.15, and 0.05. 

\textbf{FullBodyManipulation} \cite{li2023object} consists of 10 hours of high-quality, paired object and human motion, including interaction with 15 different objects. Each 3D HOI sequence pairs with a textual description used to guide volunteer acting. 17 subjects participated in the motion collection. They interact with each object following a given textual instruction. Furthermore, this dataset also selected 15 objects commonly used in everyday tasks, which include a vacuum, mop, floor lamp, clothes stand, tripod, suitcase, plastic container, wooden chair, white chair, large table, small table, large box, small box, trashcan, and monitor. We adopt the official evaluation protocol used in CHOIS \cite{li2023controllable} and OMOMO \cite{li2023object}.

\textbf{GRAB} \cite{taheri2020grab} is a large-scale whole-body grasp dataset that contains full 3D shape and pose sequences of 10 subjects interacting with 51 everyday objects of varying shape and size. The intentions of collected body-object interaction samples include “\textit{use}”, “\textit{pass}”, “\textit{lift}”, and “\textit{off-hand}”, a set of basic daily activities. Following \cite{diller2022forecasting,ghosh2022imos}, we spilt $S2 \sim S9$ subjects into the training set. The remaining $S1$ and $S10$ subjects are used for validation and testing, respectively. Thus our train, validation, and test splits respectively consist of 789, 157, and 115 body-object interaction sequences. 

\begin{table*}[t]
	\centering
	\caption{Quantitative Comparison on HIMO. Following the common-used evaluation scheme, we repeat the evaluation 20 times and report the average with 95\% confidence interval. The best and second-best results are bolded and underlined, respectively}
	\vspace{-1.5mm} 
	\label{Tab.1}
	\scalebox{0.97}{
		\begin{tabular}{lccccccc} 
			\toprule
			Method & R-Precision (Top-3) $\uparrow$ & FID $\downarrow$ & MM-Dist $\downarrow$ & Diversity $\rightarrow$ & MModality $\uparrow$ & $C_{prec}$ $\uparrow$  & $C_{\%}$ $\uparrow$ \\ \midrule
			Real & \et{0.7988}{.0081}     &  \et{0.0176}{.0065}   & \et{3.5659}{.0109}        &  \et{11.3973}{.2577}   &    ---   &  --- & ---  \\ \midrule
			IMoS \cite{ghosh2022imos}	& \et{0.5013}{.0120}      & \et{7.5890}{.1121}        &  \et{8.7402}{.0310}    & \et{7.0033}{.3205} & \et{0.9920}{.2004}          & \et{0.44}{.0021}  & \et{0.31}{.0028}   \\ 
			MDM \cite{tevet2022human}	& \et{0.6052}{.0099}      & \et{6.8457}{.3315}        &  \et{8.0187}{.0500}    & \etr{11.3891}{.2342} & \et{1.2880}{.2110}          & \et{0.49}{.0013}  & \et{0.38}{.0027}   \\
			PriorMDM \cite{shafirhuman}	& \et{0.5891}{.0031}      & \et{7.8517}{.2516}        &  \et{7.2509}{.0065}    & \et{12.5799}{.1460} & \et{1.5911}{.1449}          & \et{0.41}{.0008}  & \et{0.33}{.0022} \\
			OMOMO\cite{li2023object}&\et{0.5929}{.0124}&\et{6.1322}{.2715}&\et{7.9215}{.0659}&\et{12.7311}{.1967}&\et{1.3615}{.1667}&\et{0.43}{.0019}&\et{0.35}{.0047}  \\
			Text2HOI\cite{cha2024text2hoi}&\et{0.6033}{.0254}&\et{5.3959}{.3964}&\et{6.3697}{.4150}& \et{12.2359}{.2667} & \et{1.5449}{.1099} & \et{0.51}{.0267} & \et{0.40}{.0083} \\
			CG-HOI \cite{diller2024cg}     & \et{0.5829}{.0360} & \et{5.6961}{.3664} & \et{6.1336}{.8960} & \et{12.5996}{.3742} & \et{1.3368}{.2144} & \et{0.52}{.0029} & \et{0.48}{.0155}  \\
			CHOIS \cite{li2023controllable}& \et{0.5677}{.0411} & \et{3.9963}{.5870} & \et{5.9866}{.6931} & \et{12.4496}{.5142} & \et{1.5962}{.1987} & \etbb{0.56}{.0336} & \etbb{0.42}{.0533}  \\
			HIMO-Gen \cite{lv2024himo}     & \etbb{0.6369}{.0032}      & \etbb{1.4811}{.0427}        &  \etbb{3.6491}{.0101}    & \et{11.6603}{.2043} & \etbb{1.7863}{.0570}          & ---  & --- \\ \midrule
			EigenActor (Ours) & \etr{0.6805}{.0021}      & \etr{1.1043}{.0357}    & \etr{3.6011}{.0257}     & \etbb{11.5459}{.3052}   & \etr{1.8905}{.0234}           & \etr{0.85}{.0021}    & \etr{0.67}{.0059}  \\
			\bottomrule
	\end{tabular}}
\end{table*}

\begin{table*}[t]
	\centering
	\caption{Quantitative Comparison on FullBodyManipulation and GRAB. Following the common-used evaluation scheme, we repeat the evaluation 20 times and report the average with 95\% confidence interval. The best and second-best results are bolded and underlined, respectively}
	\vspace{-1.5mm} 
	\label{Tab.2}
	\scalebox{1}{
		\begin{tabular}{lccccccccc} 
			\toprule
			\multirow{2}{*}{Method} & \multicolumn{4}{c}{FullBodyManipulation}                &            & \multicolumn{4}{c}{GRAB}        \\  \cmidrule(lr){2-5} \cmidrule(l){7-10} 
			& R-Precision (Top-3) $\uparrow$ & FID $\downarrow$ & MM-Dist $\downarrow$ & \multicolumn{1}{c}{$C_{prec}$ $\uparrow$} &  & R-Precision (Top-3) $\uparrow$ & FID $\downarrow$ & MM-Dist $\downarrow$ & \multicolumn{1}{c}{$C_{prec}$ $\uparrow$}\\ \midrule
			Real & $0.82$     &  $0.088$   & $2.86$    &  ---   & & $0.77$     &  $0.002$   & $3.11$   &  --- \\ \midrule
			OMOMO\cite{li2023object} & $0.54$     &  $5.21$   & $6.99$    &  $0.44$   & & $0.59$     &  $4.82$   & $6.49$   &  $0.35$  \\
			Text2HOI\cite{cha2024text2hoi}& $0.60$     &  $4.99$   & $6.21$    & $0.52$   & & $0.56$     &  $3.98$   & $5.66$   &  $0.42$  \\
			CG-HOI \cite{diller2024cg}    & $0.55$     &  $5.67$   & $5.96$    &  $0.55$   & & $0.50$     &  $4.45$   & $5.35$   &  $0.45$  \\
			CHOIS \cite{li2023controllable} & $0.64$     &  $\underline{0.69}$   & $\underline{3.96}$    &  $\underline{0.80}$   & & $0.59$     &  $\underline{1.34}$   & $4.96$   &  $\underline{0.65}$  \\
			HIMO-Gen \cite{lv2024himo}   & $\underline{0.66}$     &  $1.83$   & $4.82$    &  $0.69$   & & $\underline{0.61}$     &  $2.96$   & $\underline{4.33}$   &  $0.51$   \\ \midrule
			EigenActor (Ours) & $\boldsymbol{0.73}$     &  $\boldsymbol{0.62}$   & $\boldsymbol{3.75}$    &  $\boldsymbol{0.87}$   & & $\boldsymbol{0.66}$     &  $\boldsymbol{0.89}$   & $\boldsymbol{4.07}$   &  $\boldsymbol{0.72}$  \\
			\bottomrule
	\end{tabular}}
\end{table*}

\subsection{Evaluation Metrics}
Following \cite{lv2024himo, diller2024cg, chen2023executing, jiang2023motiongpt}, we adopt five quantitative metrics to evaluate the performance of our EigenActor on text-to-HOI syntheses comprehensively, including the semantic consistency, interaction realism, and generation diversity. As proposed in \cite{lv2024himo}, a HOI feature extractor and text feature extractor used in the text-to-HOI synthesis evaluation are pre-trained in a contrastive manner. In the following, we briefly introduce these evaluation metrics. \textit{Frechet Inception Distance} (FID) evaluates the generation realism by computing the latent feature distribution distance between the generated and real HOIs; \textit{R-Precision} reflects the semantic consistency between generated HOIs and the given textual prompts via a retrieval strategy; \textit{Multi-Modal Distance} (MM Dist) reflects the relevance between generated HOIs and given textual prompts by computing the average Euclidean distance between their features; \textit{Diversity} (DIV) measures the variability of the generated HOIs by computing the variance of feature vectors of generated HOIs across all text descriptions; \textit{MultiModality} (MModality) shows the mean-variance of generated HOI conditioned on a single text prompt. 

In particular, following prior work \cite{li2023object, li2023controllable}, we further adopt other two quantitative metrics to evaluate the interaction quality of synthesized 3D HOI samples. Specifically, we first compute the distance between hand positions and object meshes. We empirically set a contact threshold (5cm) and use it to extract contact labels for each frame. Then, we perform the same calculation for ground truth hand positions. Thus, \textit{Contact Precision} ($C_{prec}$) counts true/false positive/negative cases to compute hand-object contact accuracy; \textit{Contact Percentage} ($C_{\%}$) reflects the frame-level contact inference accuracy, determined by the proportion of frames where contact is detected.

\subsection{Baseline Methods}
To comprehensively verify the effectiveness of EigenActor, we compare its performances with several strong text-to-HOI baseline methods. Specifically, similar to \cite{lv2024himo}, we first adapt MDM \cite{tevet2022human}, PriorMDM \cite{shafirhuman}, and OMOMO \cite{li2023object} to support the joint condition inputs of object geometry and text instruction. Besides, we also develop Text2HOI \cite{cha2024text2hoi} from a hand-object interaction synthesis system to full-body interaction synthesis. Furthermore, we re-train CHOIS \cite{li2023controllable} and HIMO-Gen \cite{lv2024himo} with the GRAB \cite{taheri2020grab} dataset. We also re-implement CG-HOI \cite{diller2024cg} based on its official methodology descriptions and default model configurations.

\begin{figure*}[t]
	\centering
	\includegraphics[width=1\textwidth]{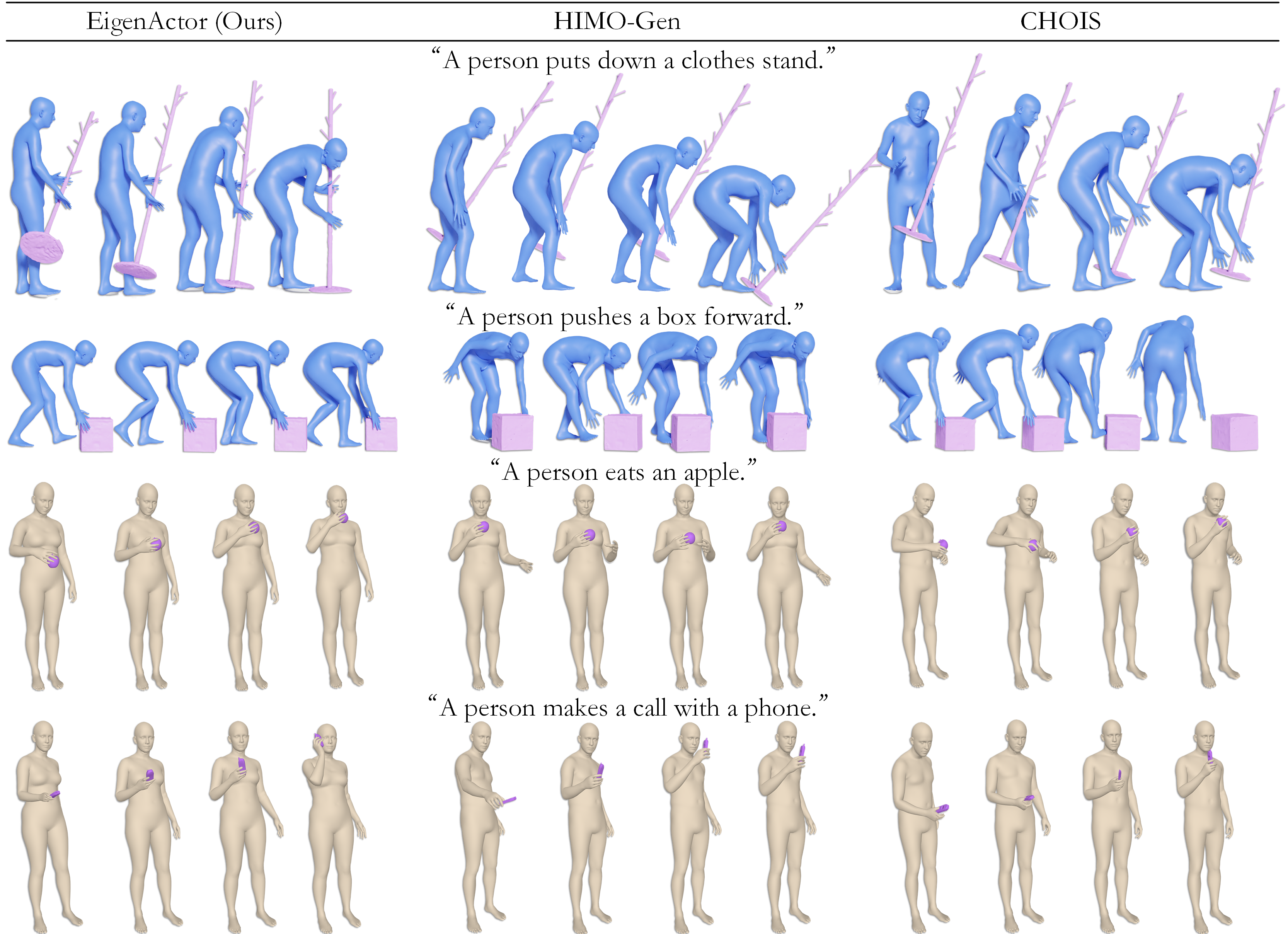}	
	\vspace{-6mm}
	\caption{Qualitative comparison between ours and state-of-the-art methods. We visualize body-object interaction samples synthesized from different given text-object conditions.Top-2 rows (blue bodies) and bottom-2 rows (brown bodies) visualize the HOI samples synthesized from the FullBodyManipulation and GRAB test sets, respectively.}
	\label{vis_compare}
\end{figure*} 

\subsection{Comparison to State-Of-The-Art Methods}
To validate the proposed EigenActor, we show extensive performance comparisons between different text-to-HOI synthesis methods on three large-scale datasets, including quantitative, qualitative, few-sample learning, and user study comparisons. In the following, we analyze these performance comparisons, respectively.
\subsubsection{Quantitative Comparisons} 
In this section, we analyze the performance of our and previous methods via their quantitative comparisons. Analyzing the text-to-HOI synthesis performance comparisons on the HIMO dataset (Tab. \ref{Tab.1}), we observe that EigenActor has several advantages: (I) Better Text-HOI Semantic Consistency. Compared with CHOIS and HIMO-Gen, EigenActor outperforms them on the Top-3 R-Precision evaluation metric with 21\% and 8\% performance gains. These performance improvements on Top-3 R-Precision indicate that the 3D HOI samples generated from EigenActor are more consistent with their given text-based instruction conditions; (II) More Realistic Body-Object Interactions. EigenActor outperforms HIMO-Gen with significant performance improvements on FID and MM-Dist evaluations. Furthermore, notable performance gains on $C_{prec}$ and $C_{\%}$ also verify that 3D HOI samples inferred from EigenActor have better hand-object contact accuracy and physical reasonableness; (III) Better Generation Diversity. Besides more realistic generation, EigenActor also outperforms CHOIS and HIMO-Gen with a better diversity performance in text-to-HOI synthesis. As shown in Tab. \ref{Tab.2}, EigenActor also has performance superiorities on FullBodyManipulation and GRAB datasets, indicating its powerful robustness and generalization on diverse large-scale datasets. All these experimental results on three datasets verify that EigenActor is a powerful text-to-HOI synthesis system with advantages in consistent text-HOI semantics, realistic body-object interaction, and diverse generation.

\begin{figure}[t]
	\centering
	\includegraphics[width=0.46\textwidth]{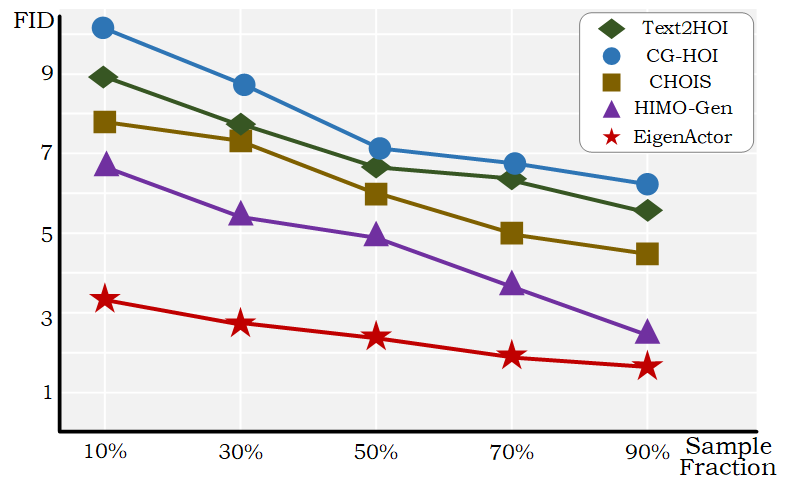}	
	\vspace{-3mm}
	\caption{Text-to-HOI synthesis performances with fewer training samples on HIMO.}
	\label{few_sample}
\end{figure}

\begin{figure}[t]
	\centering
	\includegraphics[width=0.48\textwidth]{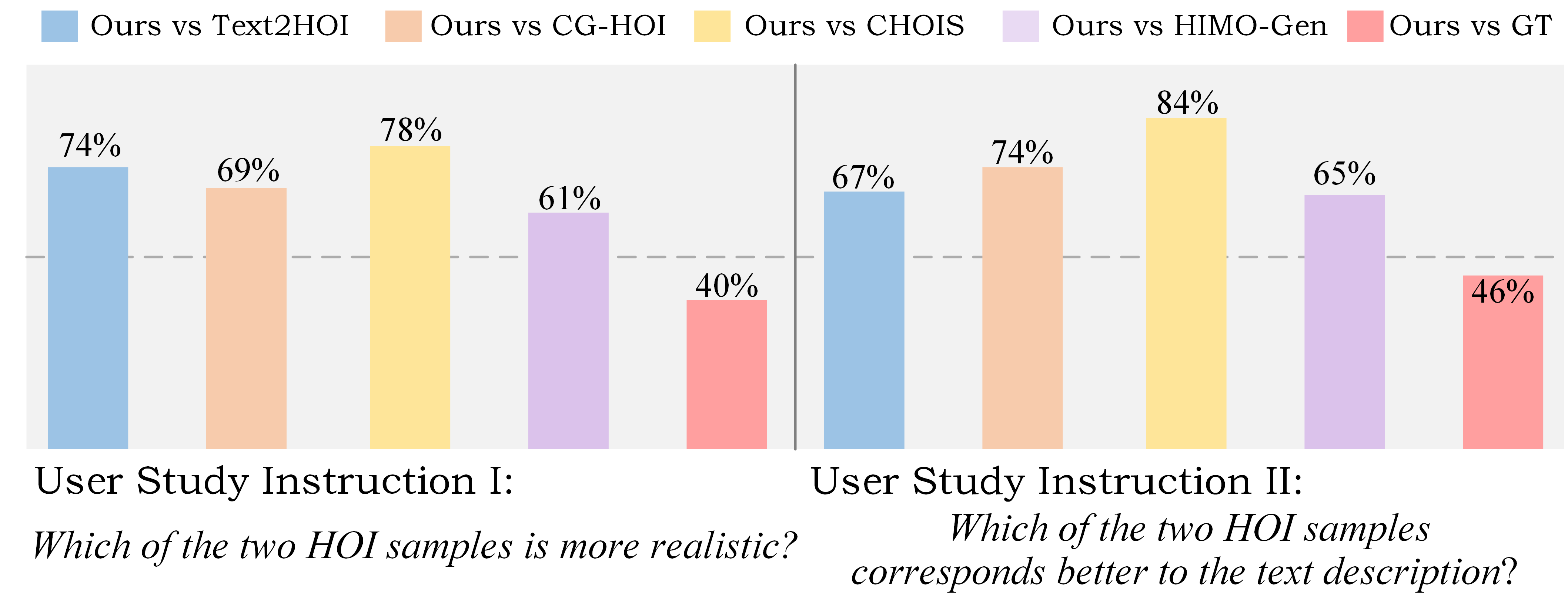}	
	\vspace{-3mm}
	\caption{User Study. Each bar indicates the preference rate of our proposed EigenActor model over other text-to-HOI synthesis methods.}
	\label{user_study}
\end{figure} 

\subsubsection{Qualitative Comparisons}
We further evaluate the performance of different text-to-HOI synthesis methods via a qualitative comparison between their synthesized samples. As shown in Fig. \ref{vis_compare}, we randomly select two object-text condition pairs from the test set of FullBodyManipulation and GRAB and respectively visualize 3D HOI samples generated from them. Analyzing the visualization comparisons shown in Fig. \ref{vis_compare}, we see that the HOI samples generated from EigenActor outperform others with better motion naturalness and interaction realism. For example, HOI samples generated from HIMO-Gen or CHOIS both suffer from poor body-object interaction realism, such as body-object inter-penetration, off-hand and pose disharmony. Besides, HIMO-Gen and CHOIS also have inconsistency problems between their poses and intended semantics and disharmony problems between body and object movements. These extensive qualitative comparisons also verify the effectiveness of the proposed EigenActor. 

\begin{figure*}[t]
	\centering
	\includegraphics[width=1\textwidth]{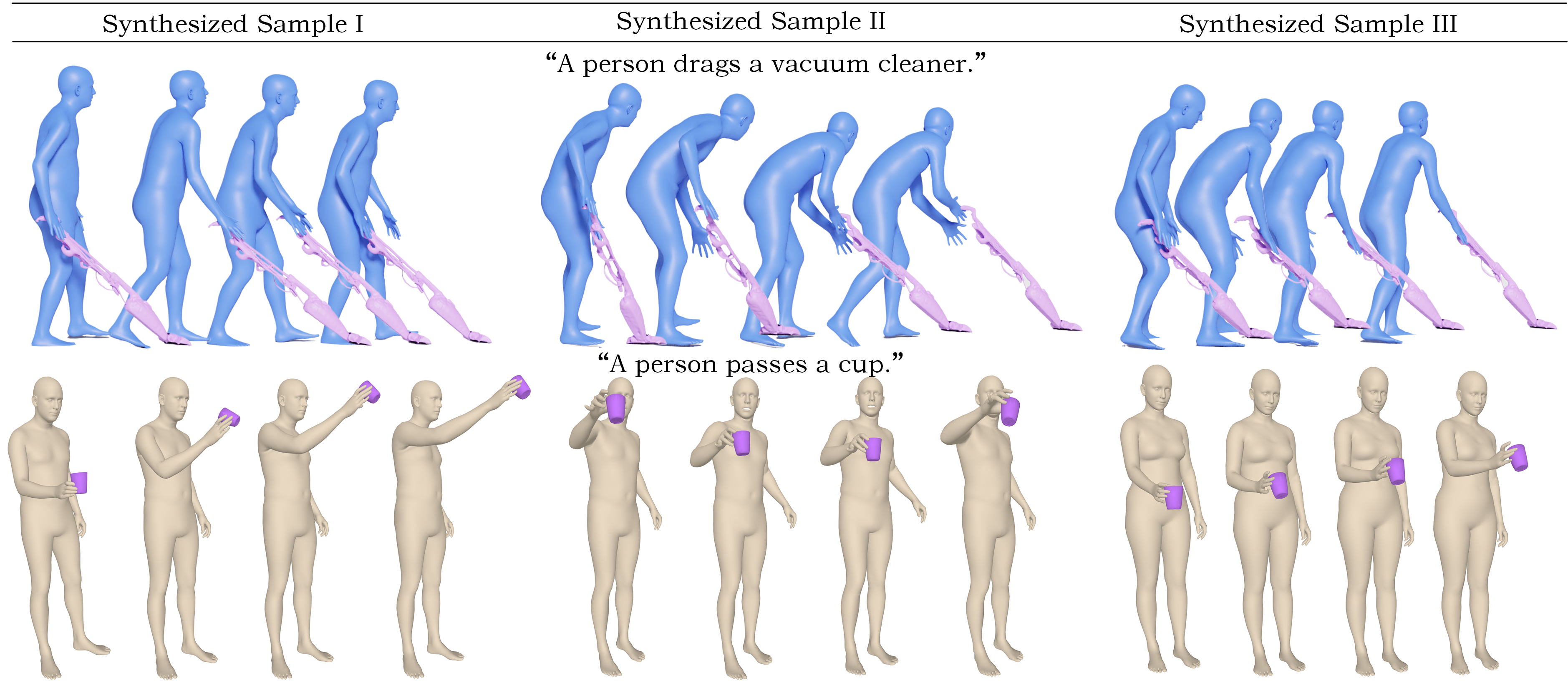}	
	\vspace{-3mm}
	\caption{Generation Diversity Visualization. We visualize diverse HOI examples synthesized from the same given text-object condition contexts.}
	\label{vis_div}
\end{figure*} 

\subsubsection{Synthesis with Fewer Samples}
To investigate the robustness of EigenActor on a limited number of paired text-HOI samples, we first retrain these text-to-HOI synthesis methods by randomly sampling a fraction of the HIMO training dataset and then evaluate their FID performances on the original test dataset. As verified in Fig. \ref{few_sample}, EigenActor significantly outperforms other baseline methods when training with 10\%, 30\%, 50\%, 70\%, and 90\% training samples. For example, when training with 10\% samples, EigenActor outperforms HIMO-Gen with 50\% performance gains on the FID metric. It verifies that EigenActor encapsulates effective action-specific priors from limited body motion samples. Benefiting from this intra-class priori knowledge, EigenActor improves body interaction synthesis with richer condition context, thus significantly facilitating few-sample cross-modality map learning.

\subsection{Perceptual User Study}
In this section, we evaluate text-to-HOI generation performances with perceptual user studies. Specifically, as shown in Fig. \ref{user_study}, we adopt a force-choice paradigm that asks “\textit{Which of the two HOI samples is more realistic}?” and “\textit{Which of the two HOI samples corresponds better to the text description}?”. These two questions focus on evaluating our performances on HOI interaction realism and text-HOI semantics consistency, respectively. The provided 3D HOI samples are generated from 30 text-object condition pair inputs randomly selected from the test set of the GRAB dataset. Then, we invite 20 subjects and provide five comparison pairs: ours vs Text2HOI, ours vs CG-HOI, ours vs CHOIS, ours vs HIMO-Gen, and ours vs Ground-Truth. As verified in Fig. \ref{user_study}, our method outperforms the other state-of-the-art methods on both HOI interaction realism and text-HOI semantics consistency performances by large margins and is even comparable to the real HOIs.

\subsection{Generation Diversity}
In this section, we visualize more synthesized HOI samples to analyze their generation diversity performances. As shown in Fig. \ref{vis_div}, we choose two different text-object pairs and consider them as the condition inputs of EigenActor. Then, we respectively visualize three different HOI samples synthesized from the same text-object condition contexts. Analyzing the visualizations shown in Fig. \ref{vis_div}, we can see that EigenActor can generate diverse and realistic 3D HOI samples with different interaction styles that both conform to their given object shape conditions and intended interaction semantics. The different HOI samples synthesized from the given “\textit{a person passes a cup}” instruction enrich intra-class interaction diversity with different hand motion paths. These realistic and diverse body-object interaction samples verify the synthesis diversity capability of EigenActor.

\begin{table}[t]
	\centering
	\caption{Comparisons between different text-to-HOI synthesis strategies with/without action-specific motion basis.}
	\vspace{-1.5mm} 
	\label{Tab.3}
	\scalebox{0.9}{
		\begin{tabular}{cc|ccc}
			\toprule
			\multicolumn{2}{c|}{Action-specific Motion Basis} & \multirow{2}{*}{\begin{tabular}[c]{@{}c@{}}R-Precision \\ (Top-3)\end{tabular} $\uparrow$} & \multirow{2}{*}{FID $\downarrow$} & \multirow{2}{*}{$C_{prec}$ $\uparrow$} \\ \cmidrule{1-2}
			Inferred                  & Real                  &                                                                             &                      &                                             \\ \midrule
			$\times$             &      $\times$                & $0.48$               & $2.31$                   & $0.41$   \\ 
			$\times$             &      $\checkmark$            & $0.69$               & $0.82$                   & $0.77$    \\
			$\checkmark$          &      $\times$               & $0.66$               & $0.89$                   &  $0.72$   \\
			
			\bottomrule
	\end{tabular}}
\end{table}

\subsection{Ablation Study}
In this section, we analyze the effectiveness of individual components and investigate their configurations in the final EigenActor architecture. Unless stated, the reported performances are Top-3 R-Precision, FID, and $C_{prec}$ metrics on GRAB dataset.

\begin{figure*}[t]
	\centering
	\includegraphics[width=1\textwidth]{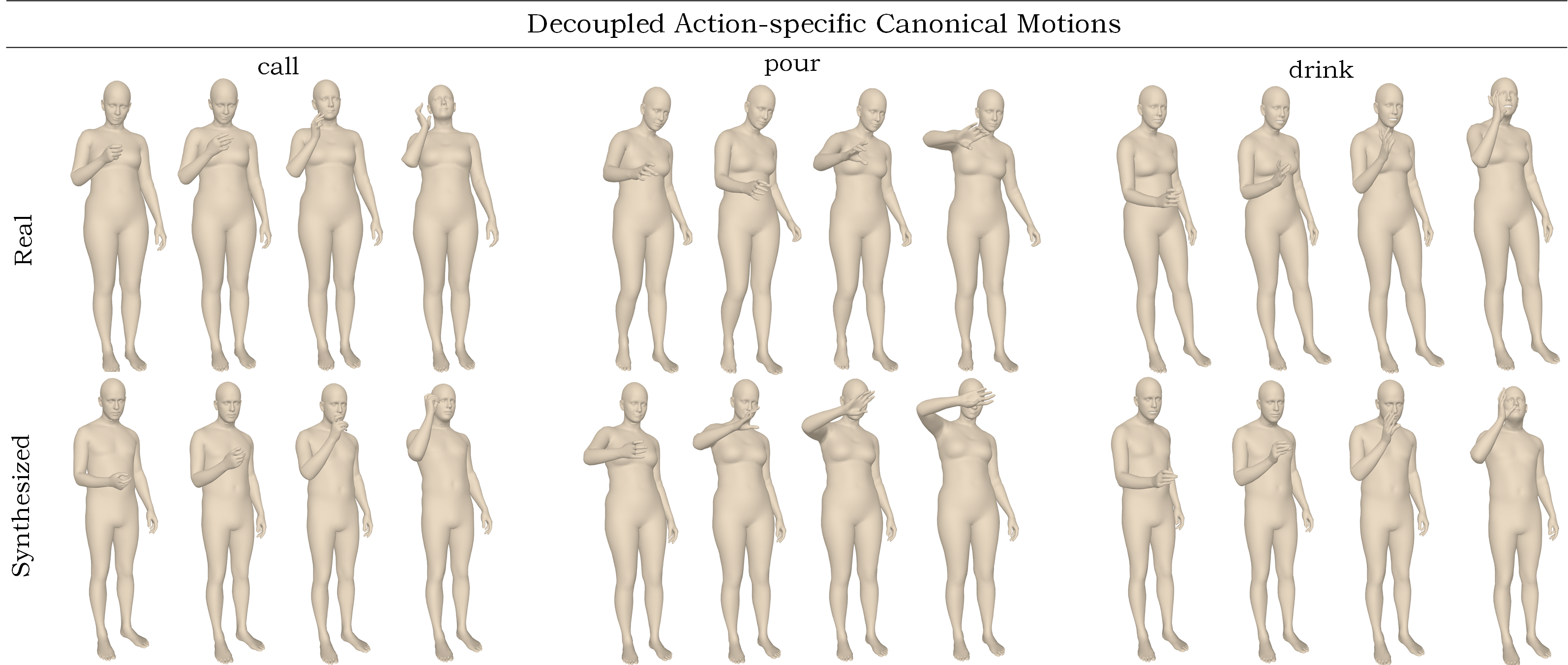}	
	\vspace{-3mm}
	\caption{Decoupled Action-specific Motion Visualization. We visualize the synthesized and real action-specific motions of three different action categories.}
	\label{eigen_motion}
\end{figure*} 

\subsubsection{Effect of Decoupling Action-specific Motion Prior}
In this section, we explore the effectiveness of decoupled action-specific motion priors via two ablative experiments. Specifically, as shown in Tab. \ref{Tab.3}, we first compare text-to-HOI synthesis performances between three different body generation strategies ($i.e.$, setup I $\sim$ III). In setup I, without the action-specific motion inference stage, we directly generate body motions from text-object conditions. In setup II and III, we adopt a factorized body motion inference scheme in their training stages but respectively use real and inferred action-specific motion basis in their test stages. As verified in Tab. \ref{Tab.3}, compared with setup I, setup II and III obtain better text-to-HOI synthesis performances, thus verifying the effectiveness of the proposed action-specific motion inference strategy. Furthermore, without introducing a real action-specific motion basis into the text-to-HOI synthesis, setup III obtains a performance comparable to setup II, validating the prediction accuracy of the inferred action-specific motions.  

We further respectively visualize the inferred action-specific motion samples and their ground truths. As shown in Fig. \ref{eigen_motion}, taking three action categories as examples, the action-specific canonical body motion sequences inferred from given text-based interaction semantics are visually consistent with their real ones. For example, in the \textit{pour} action, an inferred canonical body pose sequence and its real one characterize similar motion pattern cues ($i.e.$, flipping the wrist while lifting the arm). These visualization comparisons also verify the effectiveness of the proposed action-specific motion inference module.

\begin{figure}[t]
	\centering
	\includegraphics[width=0.48\textwidth]{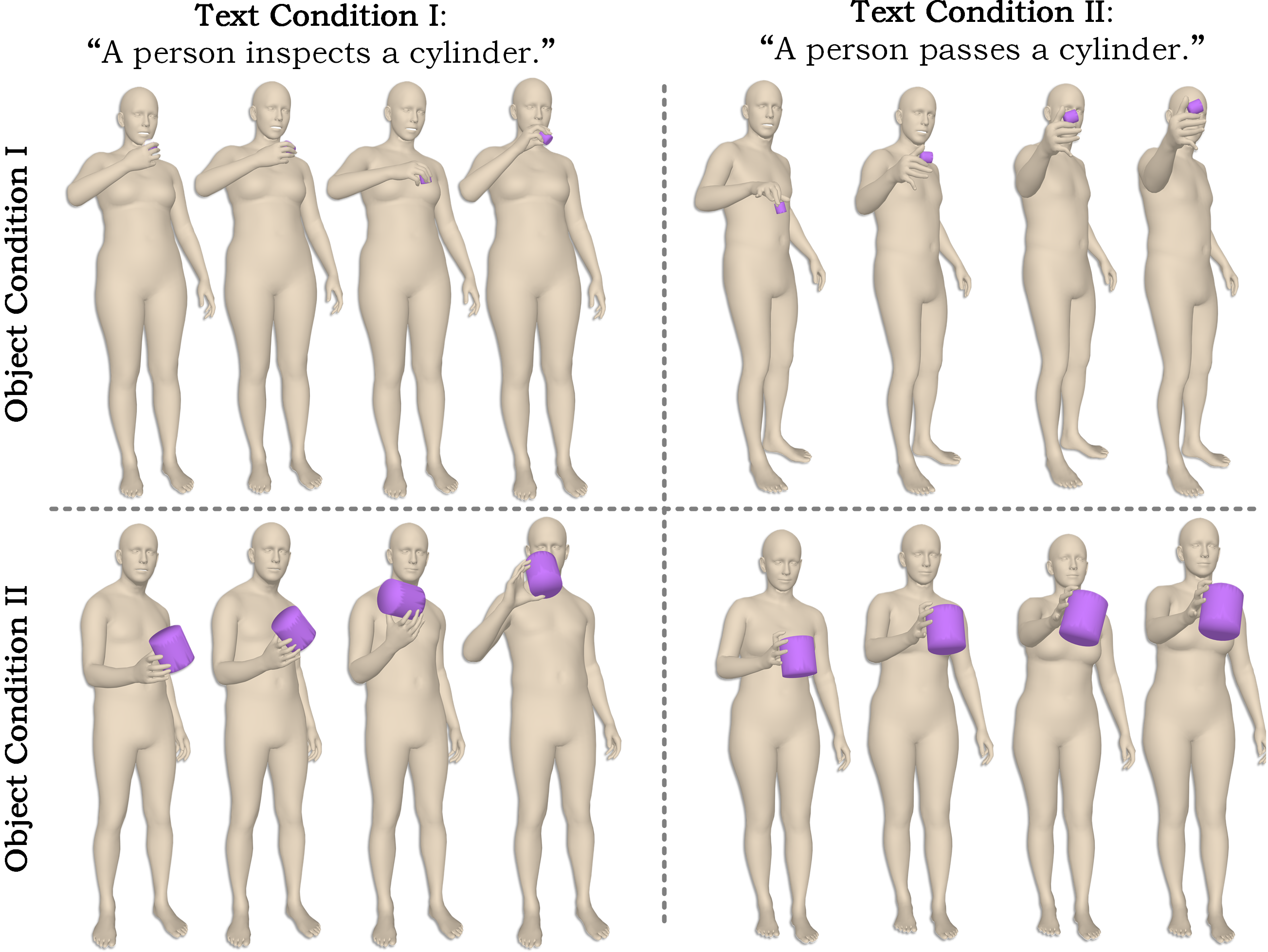}	
	\vspace{-3mm}
	\caption{We choose two text instruction conditions and pair each of them with two different object shape conditions. Then, we respectively visualize four HOI samples synthesized from these four different text-object conditions.}
	\label{t_o_conds}
\end{figure} 

\begin{table}[t]
	\centering
	\caption{Comparisons between different text-to-HOI synthesis strategies with/without object contact part inference.}
	\vspace{-1.5mm} 
	\label{Tab.object_net_ablation}
	\scalebox{0.87}{
\begin{tabular}{c|ccc}
	\toprule
	\begin{tabular}[c]{@{}c@{}}Condition inputs of the\\object-specific interaction diffusion\end{tabular} & \begin{tabular}[c]{@{}c@{}}R-Precision\\ (Top-3)\end{tabular} $\uparrow$ & FID $\downarrow$& $C_{prec}$ $\uparrow$\\ \midrule
	$f_t$-only                                                           &      $0.58$                 &  $0.95$   &  $4.32$ \\ 
	$f_o$-only                                                           &      $0.61$                 &  $0.92$   &  $4.21$ \\ 
	$f_t$ and $f_o$                                                      &      $0.66$                 &  $0.89$   &  $4.07$\\ \bottomrule
\end{tabular}
}
\end{table}

\begin{table}[t]
	\centering
	\caption{Comparisons between different text-to-HOI synthesis strategies with/without object contact part inference.}
	\vspace{-1.5mm} 
	\label{Tab.4}
	\scalebox{0.9}{
		\begin{tabular}{cc|ccc}
			\toprule
			\multicolumn{2}{c|}{$\quad$ Object Contact Parts $\qquad$} & \multirow{2}{*}{\begin{tabular}[c]{@{}c@{}}R-Precision \\ (Top-3)\end{tabular} $\uparrow$} & \multirow{2}{*}{FID $\downarrow$} & \multirow{2}{*}{$C_{prec}$ $\uparrow$} \\ \cmidrule{1-2}
			Inferred                  & Real                  &                                                                             &                      &                                             \\ \midrule
			$\times$             &      $\times$                & $0.59$               & $1.99$                   & $0.41$   \\ 
			$\times$             &      $\checkmark$            & $0.71$               & $0.84$                   & $0.80$    \\
			$\checkmark$         &      $\times$               & $0.66$               & $0.89$                   &  $0.72$   \\
			
			\bottomrule
	\end{tabular}}
\end{table}

\begin{figure}[t]
	\centering
	\includegraphics[width=0.44\textwidth]{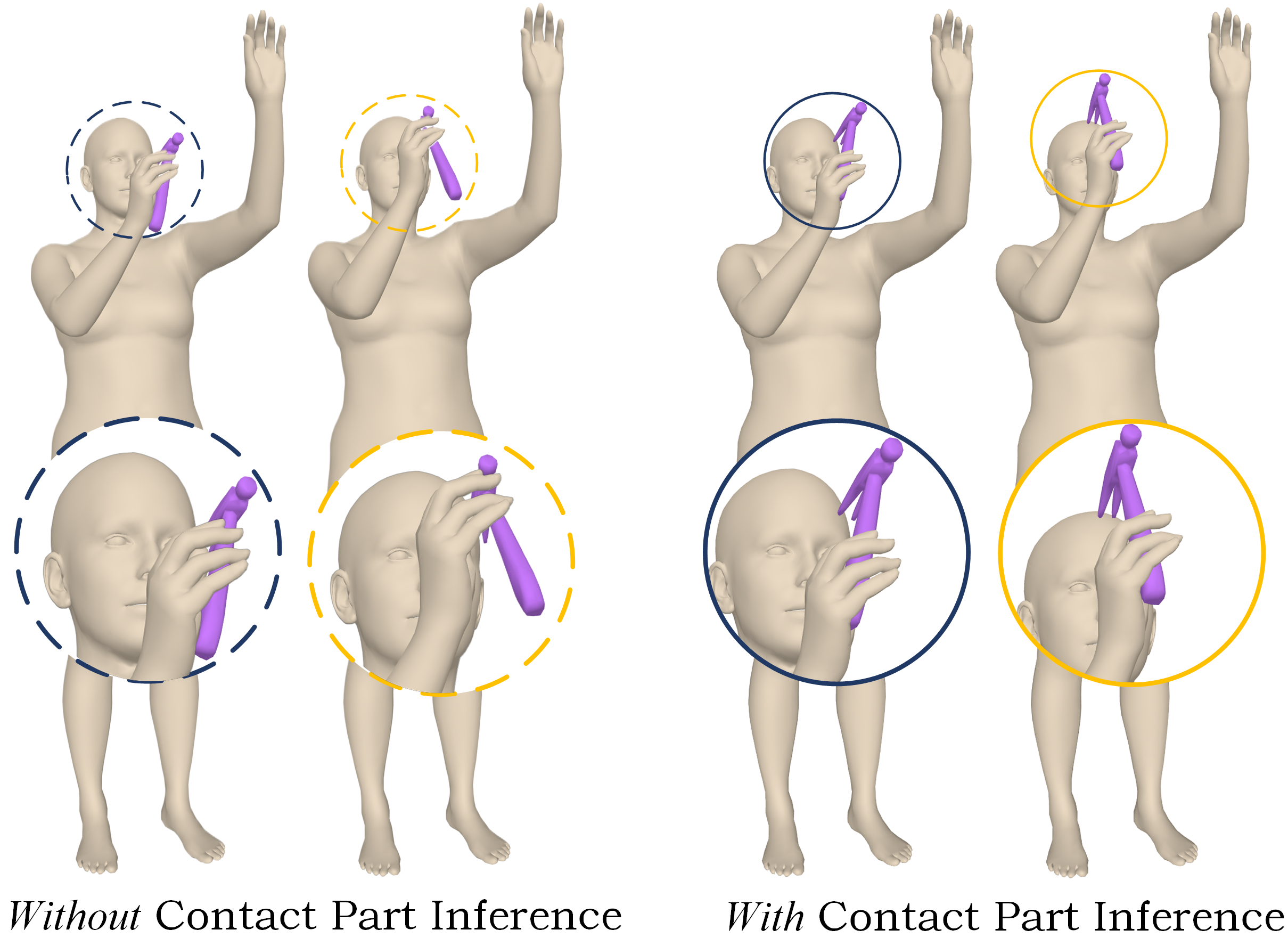}	
	\vspace{-3mm}
	\caption{Visualization comparisons between HOI samples synthesized with/without the object contact part inference module. Richer object contact part conditions benefit inferring more realistic in-hand manipulations.}
	\label{with_without_conact_infer}
\end{figure} 

\subsubsection{Effect of Decoupling Object-specific Interaction Prior}
In this section, we explore the effectiveness of the learned object-specific interaction priors. Specifically, we first choose two text instruction conditions and pair each of them with two different object shape conditions. Then, we respectively visualize four HOI samples synthesized from these four different text-object conditions. As shown in Fig. \ref{t_o_conds}, 3D body pose samples are consistent with their intended interaction semantics and also interact with their given specific objects realistically. For example, given the same \textit{inspecting} command and two different \textit{cylinder} objects, two HOI samples synthesized from these two text-object conditions both reflect the same inspecting intention while their in-hand grasp postures vary with the different object shape priors. These visualization analyses verify that EigenActor jointly understands text-object cross-modality conditions and learns effective object-specific interaction style priors. 

Besides, we also tune the condition inputs of the object-specific interaction diffusion module to explore their effects. Specifically, as shown in Tab. \ref{Tab.object_net_ablation}, compared with text-only and object-only condition inputs, introducing text-object joint inputs into the object-specific interaction inference can bring 13\% and 8\% top-3 R-Precision performance improvements, respectively. They indicate that text-based motion intention and 3D object geometries are important priors for object-specific interaction style inference.
  
\begin{figure}[t]
	\centering
	\includegraphics[width=0.48\textwidth]{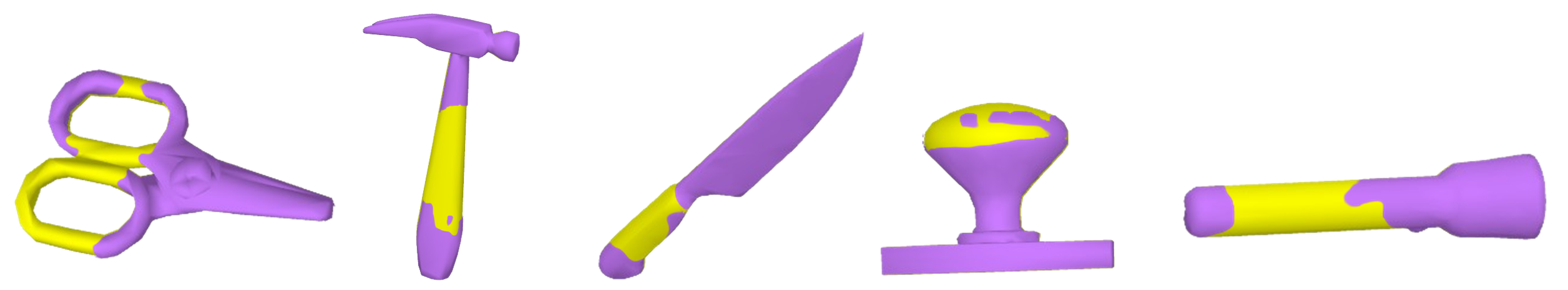}	
	\vspace{-3mm}
	\caption{Visualizations of inferred contact parts of different objects. The inferred contact parts of the object are marked in yellow.}
	\label{contact_infer}
\end{figure} 

\begin{table}[t]
	\centering
	\caption{Performance comparisons between text-to-HOI synthesis with different hand-object interaction optimization constraints.}
	\vspace{-1.5mm} 
	\label{Tab.5}
	\scalebox{0.9}{
		\begin{tabular}{cc|ccc}
			\toprule
			\multicolumn{2}{c|}{Interaction Optimization Constraints } & \multirow{2}{*}{\begin{tabular}[c]{@{}c@{}}R-Precision \\ (Top-3)\end{tabular} $\uparrow$} & \multirow{2}{*}{FID $\downarrow$} & \multirow{2}{*}{$C_{prec}$ $\uparrow$} \\ \cmidrule{1-2}
			Temporal Consistency     & In-Contact   &     &                      &             \\ \midrule
			$\times$             &      $\times$               & $0.58$               & $1.72$                   & $0.49$   \\ 
			$\times$             &      $\checkmark$           & $0.62$               & $0.95$                   & $0.61$    \\
			$\checkmark$         &      $\times$               & $0.64$               & $0.93$                   &  $0.65$   \\
			$\checkmark$         &      $\checkmark$           & $0.66$               & $0.89$                   &  $0.72$   \\
			
			\bottomrule
	\end{tabular}}
\end{table}

\subsubsection{Effect of Contact Part Inference}
To explore the effectiveness of the proposed object contact part inference module, we first compare quantitative and qualitative performances between different EigenActor configurations with/without the object contact part inference module (Tab. \ref{Tab.4} and Fig. \ref{with_without_conact_infer}). Furthermore, we also visualize the inferred contact parts of different objects (Fig. \ref{contact_infer}). Specifically, as shown in Tab. \ref{Tab.4}, we compare text-to-HOI synthesis performances between three ObjectNet variants ($i.e.$, setup I $\sim$ III). In setup I, we re-train EigenActor without the object contact inference module. In setup II and III, we introduce an object contact part inference module into EigenActor but respectively use real and inferred object contact part conditions in its test stages. As verified in Tab. \ref{Tab.4}, introducing richer object contact part conditions into the object motion planning would significantly benefit the text-to-HOI synthesis task, improving body-object interaction realism. As shown in Fig. \ref{with_without_conact_infer}, these qualitative performance comparisons also indicate that the HOI samples synthesized based on the inferred object contact part conditions perform more realistic in-hand manipulations. 

As shown in Fig. \ref{contact_infer}, we further visualize inferred contact parts of five different objects to analyze their prediction accuracy performances. Fig. \ref{contact_infer} indicates that the contact parts inferred from each object are consistent with its functional usage and affordance. For example, the contact parts inferred from hammers and knives are mainly located at their handles. These prediction results also conform to the subjective human perception, verifying the effectiveness of the proposed object contact part inference module. 

\begin{figure}[t]
	\centering
	\includegraphics[width=0.48\textwidth]{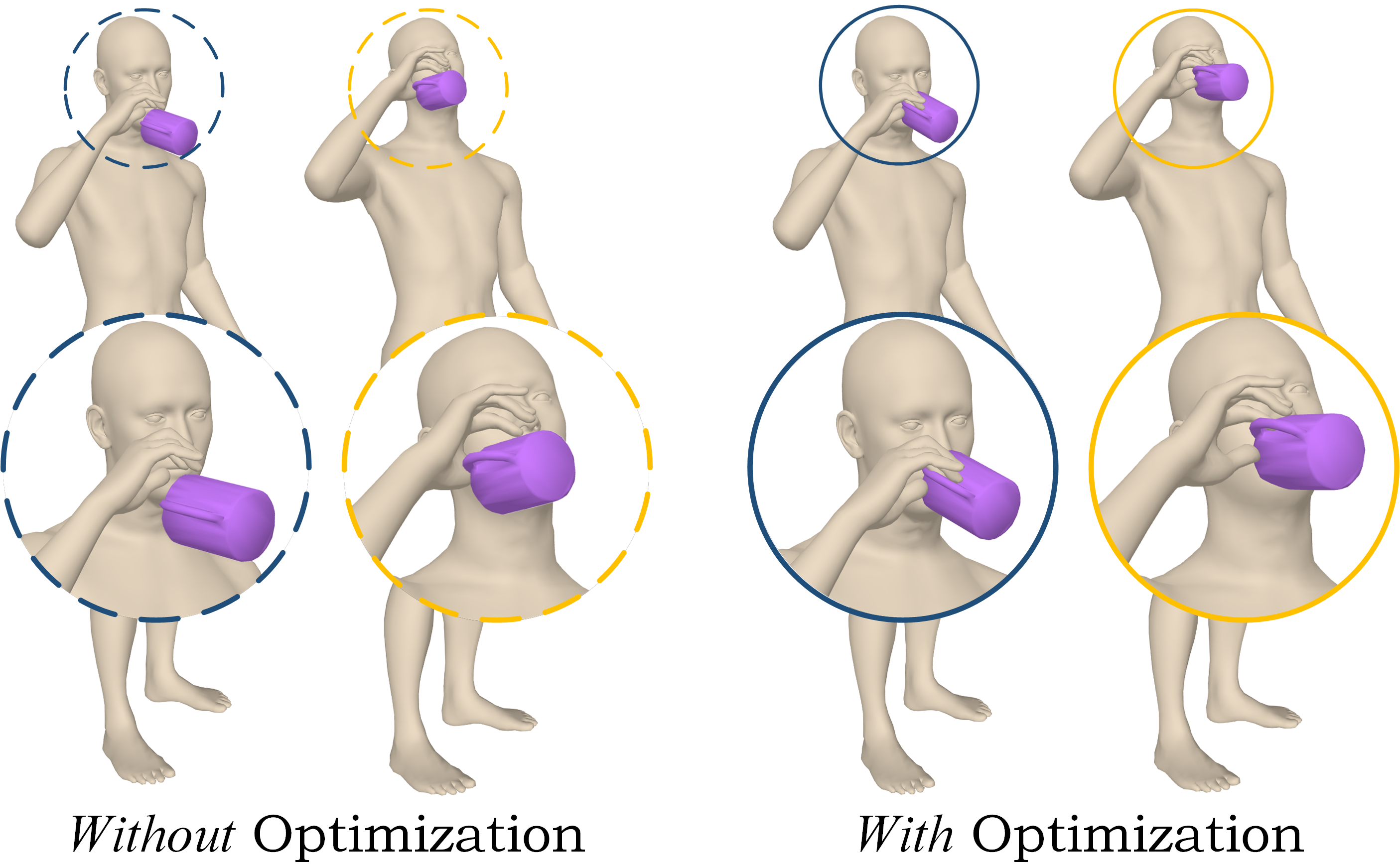}	
	\vspace{-3mm}
	\caption{Visualizations of HOI samples synthesized with/without the object-hand interaction optimization strategy.}
	\label{optimize}
\end{figure} 

\begin{table}[t]
	\centering
	\caption{Performance comparisons between different layer number and denoising step configurations.}
	\vspace{-1.5mm} 
	\label{Tab.6}
	\scalebox{0.9}{
		\begin{tabular}{c|ccc|ccc}
			\toprule
			Configurations                  & $\epsilon_\alpha$   & $\epsilon_\delta$   & $\epsilon_\gamma$   & \begin{tabular}[c]{@{}c@{}}R-Precision \\ (Top-3)\end{tabular} $\uparrow$ & FID $\downarrow$ & \begin{tabular}[c]{@{}c@{}}Time \\ (ms)\end{tabular}  $\downarrow$ \\ \midrule
			\multirow{4}{*}{Layer Number}    & 4    & 4    & 4    &   0.58      & 1.02 & 103          \\
			& 6    & 6    & 6    &   0.60      & 0.94 & 156          \\
			& 8    & 8    & 8    &   0.66      & 0.89 & 228          \\ 
			& 10   & 10   & 10   &   0.67      & 0.89 & 531          \\ \midrule
			\multirow{3}{*}{Denoising Step}  & 0.5$k$ & 0.5k & 0.5k &   0.61      & 0.92 & 141          \\
			& 1k   & 1k   & 1k   &   0.66      & 0.89 & 228        \\
			& 2k   & 2k   & 2k   &   0.67      & 0.89 & 496      \\
			& 3k   & 3k   & 3k   &   0.68      & 0.90 & 661      \\
			\bottomrule    
	\end{tabular}}
\end{table}

\subsubsection{Effect of Interaction Optimization}
We conduct extensive quantitative and qualitative performance analyses to verify the effectiveness of the hand-object interaction optimization module. Firstly, as shown in Tab. \ref{Tab.5}, we tune the constraint items deployed in the interaction optimization module and compare their performances. We can see that introducing temporal consistency and hand-object in-contact constraints into the interaction optimization module would significantly improve realistic synthesis. For example, temporal consistency and hand-object in-contact constraints bring 45\% and 44\% performance gains on the FID metric, respectively.

Furthermore, we also visualize the HOI samples generated with and without the interaction optimization module, respectively. Analyzing the qualitative comparisons shown in Fig. \ref{optimize}, we can see that the interaction optimization module introduces stronger physical constraints on in-hand interactions. Without the interaction optimization strategy, synthesized HOI samples tend to suffer from poor interaction realism, such as off-hand and misaligned contact. These quantitative and qualitative analyses both verify the effectiveness of the proposed interaction optimization module.

\subsubsection{Exploring Parameter Configurations}
In this section, we explore the optimal configurations of EigenActor, including the layer number and denoising step number of its denoisers ($i.e.$, $\epsilon_{\alpha}$, $\epsilon_{\beta}$, and $\epsilon_{\gamma}$). Specifically, we first tune the layer number of denoisers from $4$ to $10$. Then, we also provide $4$ different number choices ($i.e.$, $0.5$k, $1$k, $2$k, and $3$k) for their denoising steps. Analyzing the results shown in Tab. \ref{Tab.6}, we can see that larger denoisers and more denoising steps tend to bring better HOI synthesis performances. However, considering the brought additional computational costs, these performance gains are limited. Therefore, to balance model realism and efficiency performances, we choice $8$-layer and $1$k-step denoisers as our final model configurations.

\section{Limitation and Future Work}
In this section, we briefly analyze the limitations of EigenActor to inspire its future developments. Currently, EigenActor infers body-object 3D co-movements from given text-object joint conditions. However, because of the ambiguous nature of the linguistic descriptions, HOI samples generated from text-based instructions suffer from poor controllability on their object motion trajectories. In some real-world applications, users desire the object in generated HOI samples to move as intended. Therefore, to improve the controllability and realism of HOI syntheses, we would like to introduce planned 3D positional priors of objects and object-floor collision constraints into EigenActor.

\section{Conclusion}
In this paper, we propose a powerful text-to-HOI synthesis system named EigenActor that explores decoupled action-specific motion priors and object-specific interaction priors from HOI samples. Benefiting from the proposed two-stage body pose reasoning strategy, the body poses generated from given text-object conditions would not only conform to the intended semantics but also naturally interact with the target object. Extensive quantitative and qualitative evaluations on three large-scale datasets verify that EigenActor significantly outperforms existing SoTA methods on three core aspects: consistency between text-HOI semantics, realism of body-object interactions, and robustness of few-shot learning.

\bibliographystyle{IEEEtran}
\bibliography{egbib}

\end{document}